\documentclass[10pt]{article}
\usepackage{amssymb,amsmath}
\usepackage{mathrsfs}
\usepackage{hyperref}
\usepackage{changes}
\usepackage{algorithm}
\usepackage{amsthm}
\usepackage[noend]{algpseudocode}
\usepackage{xcolor}
\usepackage{booktabs}
\usepackage{enumerate}
\usepackage{amssymb}
\usepackage{graphicx}
\usepackage{subcaption}
\usepackage{cite}

\usepackage{tikz}
\usetikzlibrary{positioning,calc}

\newcommand{\inclu}[0] {\ar@{^{(}->}}

\newcommand{\R}{\mathbb{R}}

\newcommand{\cX}{\mathcal{X}}

\newcommand{\dom}{\text{dom}}
\newcommand{\argmin}{\operatornamewithlimits{argmin}}

\makeatletter

\newcommand{\Rmnum}[1]{\expandafter\@slowromancap\romannumeral #1@}
\makeatother

\newtheorem{theorem}{Theorem}
\newtheorem{lemma}{Lemma}
\newtheorem{assumption}{Assumption}
\newtheorem{definition}{Definition}
\newtheorem{remark}{Remark}

\definecolor{SolarizedGreen}{RGB}{133,153,0}
\definecolor{DarkBlue}{RGB}{0,0,200}
\definecolor{DarkRed}{RGB}{225,0,0}

\definecolor{codeblue}{RGB}{0,0,255}
\definecolor{codegreen}{RGB}{0,128,0}

\usepackage[listings,breakable,skins]{tcolorbox}
\newtcblisting{tcbcodeblock}[1]{%
	enhanced,
	blanker,
	top=1mm,
	bottom=1mm,
	borderline horizontal={.5pt}{0pt}{codeblue},
	colframe=codeblue,
	listing only,
	listing options={%
		style=tcblatex,
		language={#1},
		showspaces=false,
		showstringspaces=false,
		commentstyle=\color{codegreen},
		basicstyle=\ttfamily\footnotesize\color{codeblue}
	}
}

\title{Nonconvex Penalized LAD Estimation in Partial Linear Models with DNNs: Asymptotic Analysis and Proximal Algorithms}
\author{Lechen Feng%
\thanks{Department of Applied Mathematics, The Hong Kong Polytechnic University, Hong Kong. Email: {\tt fenglechen0326@163.com}}
\and Haoran Li%
\thanks{Department of Applied Mathematics, The Hong Kong Polytechnic University, Hong Kong. Email: {\tt hao316.li@connect.polyu.hk}}
\and Lucky Li%
\thanks{College of Computing, Data Science, and Society, University of California, Berkeley, CA 94720. Email: {\tt luckyql@berkeley.edu, luckyql17@gmail.com}}
\and Xingqiu Zhao%
\thanks{Department of Applied Mathematics, The Hong Kong Polytechnic University, Hong Kong. Email: {\tt xingqiu.zhao@polyu.edu.hk}}
}

\date{}

\allowdisplaybreaks[4]
\begin{document}
	\maketitle
\begin{abstract}
This paper investigates the partial linear model by Least Absolute Deviation (LAD) regression. We parameterize the nonparametric term using Deep Neural Networks (DNNs) and formulate a penalized LAD problem for estimation. Specifically, our model exhibits the following challenges. First, the regularization term can be nonconvex and nonsmooth, necessitating the introduction of infinite dimensional variational analysis and nonsmooth analysis into the asymptotic normality discussion. Second, our network must expand (in width, sparsity level and depth) as more samples are observed, thereby introducing additional difficulties for theoretical analysis. Third, the oracle of the proposed estimator is itself defined through a ultra high-dimensional, nonconvex, and discontinuous optimization problem, which already entails substantial computational and theoretical challenges.  Under such the challenges, we establish the consistency, convergence rate, and asymptotic normality of the estimator. Furthermore, we analyze the oracle problem itself and its continuous relaxation. We study the convergence of a proximal subgradient method for both formulations, highlighting their structural differences lead to distinct computational subproblems along the iterations. In particular, the relaxed formulation admits significantly cheaper proximal updates, reflecting an inherent trade-off between statistical accuracy and computational tractability.

\end{abstract}

\section{Introduction}

Partial Linear Models (PLMs) have been extensively studied in classical multivariate regression; see H{a}rdle \textit{et al.} (2006) \cite{hardle2006statistical} for comprehensive survey of this framework. The main motivation is to allow different covariates to be modeled in different ways: through simple linear effects, or through more flexible nonparametric components. In general, PLMs achieve a balance between flexibility and robustness, retaining the adaptability of nonparametric methods while reducing the dimensionality burden of fully nonparametric models. Building on these advantages, PLMs have found important applications in biostatistics, computational public health, life sciences,  environmental science, and economics; see Engle \textit{et al.} (1986) \cite{Engle01061986}, Zeger and Diggle (1994) \cite{Zegar-1994} and Peng \textit{et al.} (2006) \cite{Peng-2006}.

In this paper, we consider the following PLM:
\begin{equation}\label{plm}
Y = \beta^{\top}_0 X + g_0(Z) + \varepsilon
\end{equation}
with covariates $X \in \mathbb{R}^d$, $Z \in \mathbb{R}^l$, a vector of  unknown parameters $\beta_0$, an unknown nonlinear function $g_0$ and a random error $\varepsilon$.
Consider $N$ i.i.d. observations $\{ U_i\doteq(X_i, Y_i, Z_i) \}_{i=1}^N$, while denote $\mathbf{X}\doteq (X_1,\dots,X_N)$, $\mathbf{Y}\doteq (Y_1,\dots,Y_N)$, $\mathbf{Z} \doteq (Z_1,\dots, Z_N)$, and $\mathbf{U}\doteq(U_1,\dots, U_N)$. We aim to solve the following penalized Least Absolute Deviation (LAD) regression problem for the estimation of unknown parameters $\beta_0$ and $g_0$:
\begin{equation}\label{opt0}
	(\hat{\beta}_N,\hat{g}_N)\in\argmin_{\beta, g} \frac{1}{N} \sum_{i=1}^{N} \left|Y_i - \beta^\top X_i - g(Z_i)\right|+\lambda_N \mathcal{J}_{N,M}(\beta,g)
\end{equation}
with given $M>0$, $\lambda_N>0$ and (possibly) nonconvex and nonsmooth regularization term  $\mathcal{J}_{N,M}(\beta,g)$ bounded by constant $M$, i.e., $\|\mathcal{J}_{N,M}\|_{\infty}<M$. In this paper, we estimate the unknown function $g_0$ through a sparse Deep Neural Network (DNN), and therefore focus on the following finite-dimensional optimization problem:
\begin{equation}\label{opt0-DNN}
	\min_{\beta, g(\mathbf{W})} \frac{1}{N} \sum_{i=1}^{N} \left|Y_i - \beta^\top X_i - g(\mathbf{W};Z_i)\right|+\lambda_N \mathcal{J}_{N,M}(\beta,g(\mathbf{W})),
\end{equation}
where $g(\mathbf{W})$ belongs to a prescribed class of sparse DNNs; the readers may refer to Section \ref{sec3} for the details.

Least Squares Estimation (LSE) has been the most widely studied and influential for PLMs \eqref{plm}, due to its simplicity and broad applicability. Specifically, LSE seeks to solve the following optimization problem:
\begin{equation}\label{opt-lse}
	\min_{\beta, g} \frac{1}{N} \sum_{i=1}^{N} \left(Y_i - \beta^\top X_i - g(Z_i)\right)^2+\mathcal{J}(\beta, g)
\end{equation}
with  various parametric policies of nonlinear function $g$ and different choices of regularization term $\mathcal{J}$, as implemented in the respective literature.
Broadly speaking, methods for handling the nonparametric function $g_0$
fall into two main classes: estimating the linear and nonlinear components jointly, and disentangling the estimation of two components.
The joint estimation approach primarily relies on smoothing techniques, e.g., cubic spline smoother (Engle \textit{et al.} (1986) \cite{Engle01061986}), local polynomial smoother (Hamilton and Truong (1997) \cite{hamilton1997local}) and B-splines
smoother (McLean \textit{et al.} (2014) \cite{mclean2014functional}).
Whereas the separate estimation approach mainly includes the profile likelihood method (Severini and Wong (1992) \cite{severini1992profile}), the partial residual approach (Cuzick (1992) \cite{cuzick1992efficient},  Ferraccioli \textit{et al.} (2023) \cite{ferraccioli2023nonparametric}), and the difference approach (Duran \textit{et al.} (2012) \cite{duran2012difference}).
Meanwhile, motivated by considerations such as sparsity, smoothness, robustness, and prevention of overfitting, a line of research has focused on the selection of regularization terms $\mathcal{J}$ of problem \eqref{opt-lse}.
Henceforth, various $\mathcal{J}$ has been proposed, including Lasso (Tibshirani (1996) \cite{tibshirani1996regression}), SCAD (Smoothly Clipped Absolute Deviation, Fan and Li (2001) \cite{fan2001variable}),  Elastic Net (Zou and Hastie (2005) \cite{zou2005regularization}), MCP (Minimax Concave Penalty, Zhang (2010) \cite{zhang2010nearly}) and SACR (Smoothly Adaptively Centered Ridge, Belli (2022) \cite{belli2022smoothly}).
In recent year, with the rapid growth of big data, data have become more diverse and voluminous, bringing the new challenges for LSE. Roozbeh and Arashi (2016) \cite{roozbeh2016new} introduce a biased estimator for shrinkage parameter which is of
harmonic type mean of ridge estimators, aiming to tackle the problem of multicollinearity.
In addition, Auerbach (2022) \cite{auerbach2022identification} introduces a matching pairs  method to incorporate network data into econometric modeling.

Despite the aforementioned numerous efforts to improve LSE, it remains inherently sensitive to outliers, heavy-tailed errors, and  high reliance on assumptions (e.g., linearity and homoscedasticity) cannot be fully overcome; see {{C}}{i}{{z}}ek and Sad{i}ko{{g}}lu (2020) \cite{vcivzek2020robust} for the detailed discussion of the limitations of LSE. To circumvent the aforementioned drawbacks, LAD estimation has been adopted as a robust alternative for analyzing PLMs, i.e., estimating the unknown parameters $\beta_0$ and $g_0$ by solving optimization problem \eqref{opt0}.
Since the LAD cost function is inherently non-differentiable, even ignoring the potential non-smoothness of the regularization term $\mathcal{J}_{N,M}$, its theoretical analysis is challenging. In the early literature, to render the problem analytically tractable, unknown function $g_0$ is often represented through a basis expansion
\begin{equation*}
  g_0(Z)=\sum_{k=1}^{K} \theta_k \phi_k(Z),
\end{equation*}
where the basis functions $\{\phi_k\}_{k=1}^K$ are pre-specified, and identifying the coefficient vector $\theta = (\theta_1,\ldots,\theta_K)^\top$ is then equivalent to estimating $g_0$ itself; see He and Shi (1994) \cite{he1994convergence} and Lee (2003) \cite{lee2003efficient}.
Remarkably, such early works primarily focus on establishing the consistency and asymptotic distribution of the parametric component~$\beta_0$, rather than fully characterizing the nonparametric part.
In more recent years, to ensure the consistency of the estimation of $g_0$, alternative structural conditions are additionally required. For instance, Lian (2012) \cite{lian2012semiparametric} considers the following additive PLMs
\begin{equation*}
Y = \beta^{\top}_0 X + \sum_{k=1}^l g_{0,k}(Z_{(k)}) + \varepsilon
\end{equation*}
with $Z=(Z_{(1)},\dots,Z_{(l)})^\top$, while Ben and Lan (2016) \cite{Ben-2016AOS} further extend Lian's framework to the ultra high-dimensional setting. For a comprehensive introduction of LAD for PLMs, we refer to the monographs by Koenker \textit{et al.} (2017) \cite{regression2017handbook}.

Over the past decade, deep learning has been widely applied in many domains and has achieved remarkable success, thereby being naturally incorporated into  the traditional statistical field. Generally speaking, DNNs not only exhibit strong function approximation capabilities (see Hornik \textit{et al.} (1989) \cite{hornik1989multilayer} for the universal approximation theorem) but also help mitigate the curse of dimensionality, making them a valuable tool for estimating the nonlinear function $g_0$ of PLMs \eqref{plm}.
For instance, Farrell \textit{et al.} (2019) \cite{farrell2018deep} apply deep learning to semiparametric inference and establish nonasymptotic bounds of DNNs for nonparametric term, covering the standard LSE in particular.
Additionally, Zhong and Wang (2024) \cite{zhong2024neural} leverage deep learning for PLMs in quantile regression to achieve interpretable results and enable statistical inference, while they later extended these results to partially linear Cox models; see Zhong \textit{et al.} (2022) \cite{zhong2022deep}.  Subsequently, deep learning for PLMs via quantile regression has been extended in multiple directions, including high-dimensional PLMs (Wang (2025) \cite{wang2025partially}) and dependent data PLMs (see Brown (2024) \cite{brown2024inference} for stationary $\beta$-mixing sequences).
In a different direction, Wen \textit{et al.} (2016) \cite{NIPS2016_41bfd20a} introduce
sparse DNNs with ReLU activation function to fit unknown nonlinear function $g_0$, yielding  saving computational resources and mitigating overfitting.  Further,
Schmidt-Hieber (2020) \cite{schmidt2020nonparametric} establishes several non-asymptotic properties of the DNNs with the aforementioned sparse structure, including upper bounds on covering numbers and approximation rates for H\"older smooth functions, which provides essential theoretical tools for this paper.
In this paper, we adopt the DNN architecture of Wen \textit{et al.} (2016) \cite{NIPS2016_41bfd20a}, while allowing it to expand in width, sparsity level, and depth as the sample size increases. For notational convenience in this section, we write $\mathcal{M}_N$ for the DNN architecture associated with $N$ samples.

In conclusion, the shortcomings of the aforementioned works are as follows:
\begin{itemize}
	\item Methods based on LSE (e.g., \cite{farrell2018deep,Engle01061986,hamilton1997local,mclean2014functional,severini1992profile,cuzick1992efficient,ferraccioli2023nonparametric,duran2012difference}) are so sensitive to outliers that a single outlier can lead to completely unreliable estimates (see Hubert and Ronchetti (2009) \cite{huber1981robust} for details).
	
	\item The existing methods of estimating unknown functions, e.g. \cite{farrell2018deep,zhong2024neural,wang2025partially}, often let the regularization term exhibit very simple form (or even omit it).  However,  the estimator is usually apriori assumed to be sparse, flat, smooth and so on, leading to the nonconvex and nonsmooth regularization term, which is beyond the scope of the existing theoretical framework.
	
	\item Although the existing works such as \cite{zhong2024neural,zhong2022deep,wang2025partially} assume that the DNN architecture $\mathcal{M}_N$ expands (in width, depth, and sparsity level) as the sample size increases, the proofs of consistency and asymptotic normality rely on the fixed network architecture. This creates an inconsistency of the existing theoretical framework.
	
	\item Methods assuming additivity of the nonlinear term $g_0$ are not amenable to modeling the interaction among covariates (e.g., \cite{he1994convergence,lee2003efficient,lian2012semiparametric,Ben-2016AOS,regression2017handbook}).
	
	\item Nonparametric methods (e.g., \cite{romano2019conformalized,gan2018embedding,hatalis2017smooth}) do not leverage the known linear structure of $\beta_0^\top X$. As a result, such methods require many unnecessary parameters to approximate $\beta_0^\top X$, which can lead to the curse of dimensionality, especially when the dimension of $X$ is high.
\end{itemize}
In this paper, we propose estimator \eqref{opt0} to address the above issues. Concretely, the contributions are as follows.
\begin{itemize}
	\item
	We establish the consistency, convergence rate and asymptotic normality of estimator \eqref{opt0}. Notably, the nonconvex and nonsmooth regularization term of \eqref{opt0} invalidates the use of classical differential calculus (e.g., the chain and sum rules) on the penalized LAD criterion. This is a critical issue because a key step in establishing asymptotic normality for M-estimators relies on analyzing the differential properties of the objective function; see \cite{zhong2024neural}. Hence, we need to demonstrate that aforementioned regularization term exhibit the chain rule, additive properties and the projection theorem of partial limiting subgradient, necessitating the tools from infinite-dimensional variational and nonsmooth analysis (e.g. Mordukhovich subgradient, epi-convergence and generalized cone).
	
\item The expansion (in width, depth and sparsity level) of $\mathcal{M}_N$ architecture causes the covering number of candidate estimators to approach infinity, rendering the classic methodology (for proving consistency and asymptotic normality) inapplicable; see \cite{zhong2024neural, wang2025partially, Vaart2023} for details. To address this issue, we characterize the growth rate of the covering number and the entropy of the candidate estimators, and demonstrate the universal convergence of the criterion function \eqref{opt0}.

\item The oracle of estimator \eqref{opt0} is equivalent to a nonconvex and discontinuous optimization problem. A significant computational challenge arises when using proximal gradient-type methods, as the computational cost mainly depends on the projection operator. To balance computational tractability and precision, we propose two approaches. First, for the primal formulation, we derive a closed-form solution for the projection onto the sparse constraint. This result allows us to directly analyze the computational complexity of the proximal algorithm. Second, we approximates the primal optimization problem with a sequence of coordinate convex relaxation problems, and prove that such approximated problem converges to the primal problem. The relaxed formulation admits significantly cheaper proximal updates, reflecting an inherent trade-off between statistical accuracy and computational tractability.

\item To evaluate the optimization error, we establish the global convergence of the proximal stochastic subgradient method for both the primal and approximate formulations of the penalized LAD regression problem \eqref{opt0-DNN}. Our proof leverages the Lyapunov framework developed by \cite{davis2020stochastic,bolte2007clarke,duchi2018stochastic}, and the core of our analysis is to show the Weak Sard Property.
To prove this property, we employ tools from differential and algebraic geometry, including Whitney stratification, Sard's theorem, and the chain rule for locally Lipschitz functions.
To establish the Weak Sard Property, we partition the feasible set into a collection of disjoint smooth manifolds, and the penalized LAD cost is smooth on each manifold. We then use the classical gradient of local mollifier to cover the Clarke subgradient of the penalized LAD cost.  By applying the standard Sard's theorem to such the local mollifiers on each piece of the partition, the Weak Sard Property is deduced.

\end{itemize}

\section{Preliminaries}
\textbf{Notation:}
For \(A \in \mathbb{R}^{m \times n}\), we define
$
\sigma_{\min}(A) := \min_{\|x\|_2=1} \| A x \|_2
$.
The graph of \( f \), denoted by \( \mathbf{graph}(f) \), is defined as
$
\mathbf{graph}(f) := \left\{ (x, f(x)) \in \mathbb{R}^n \times \mathbb{R}^m \;\middle|\; x \in \mathrm{dom}(f) \right\}
$. For $n\geq 1$, the set $[n]$ denotes $\{1,\dots,n\}$.
$S$ be a subset of a topological space $X$. A point $x$ is a cluster point of the set $S$ if every neighbourhood of $x$ contains infinitely many points of $S$ different from $x$ itself. Let $\mathbf{Cluster}(S)$ denote the set of all the cluster points of $S$.
	For a set $A$, let $\|A\|:= \sup_{a\in A}\|a\|$.
Given a set of functions $\mathcal{F}$, we define $
\|G\|_{\mathcal{F}} = \sup_{f \in \mathcal{F}} | G(f) |$.
 For a sub-Gaussian random variable $X$, its $\psi_2$-norm is defined as
$
\|X\|_{\psi_2} =\inf\{ t > 0 : \mathbb{E}[\exp({X^2}/{t^2})] \le 2\}
$. Let $P$ denote the true distribution of the observations and $\mathbb{P}_n$ the empirical measure based on a sample $X_1, \dots, X_n$, that is,
$\mathbb{P}_n f = \frac{1}{n} \sum_{i=1}^n f(X_i)$ and  $P f = \mathbb{E}_P[f(X)]$.
The empirical process $\mathbb{G}_n$ is then defined by
$
\mathbb{G}_n f =\sqrt{n}(\mathbb{P}_n - P)f
$.
For the empirical measure $\mathbb{P}_n = \frac{1}{n}\sum_{i=1}^n \delta_{X_i}$,
the empirical $L^p$ space is defined as
\[
L^p(\mathbb{P}_n)
\;=\;
\Bigl\{ f \text{ measurable} :
\|f\|_{L^p(\mathbb{P}_n)}
= \Bigl( \int |f|^p \, d\mathbb{P}_n \Bigr)^{1/p}
= \Bigl( \frac{1}{n} \sum_{i=1}^n |f(X_i)|^p \Bigr)^{1/p}
< \infty \Bigr\}.
\]
We use the soft O-notation $\widetilde{O}(\cdot)$ to suppress polylogarithmic factors in complexity bounds.
Formally, $f(n) = \widetilde{O}(g(n))$ if there exists a constant $k > 0$ such that
$f(n) = O\!\left(g(n) \log^{k} n\right)$. The convex hull of the set $C$,
defined as
\[
\mathbf{conv}(C)
:= \left\{ \sum_{i=1}^k \lambda_i x_i \ \middle|\
x_i \in C,\ \lambda_i \ge 0,\ \sum_{i=1}^k \lambda_i = 1,\ k \in \mathbb{N} \right\}.
\]We use the notation \(f(x) \lesssim g(x)\) to mean that
there exists a constant \(C > 0\), independent of the relevant variables, such that
$
f(x) \le C  g(x)
$.
Given a probability space $(\Omega, \mathcal{F},P)$, define
\begin{itemize}
	\item $P^*(A)$: Outer measure of a set $A \subseteq \Omega$,
	defined as ${P}^*(A) := \inf\{P(B) \mid B \in \mathcal{A},\, A \subseteq B\}$.
	\item $\mathbb{E}^*[X]$: Outer expectation of a function $X:\Omega\to\mathbb{R}$,
	defined as $\displaystyle \mathbb{E}^*[X] := \inf\left\{ \mathbb{E}[Y] \ \middle|\ Y \text{ measurable},\ Y \ge X \right\}$.
\end{itemize}
Let $L^2(\mathbf{m})$ be the space of square-integrable functions with respect to Lebesgue measure $\mathbf{m}$. The identity map on $L^2(\mathbf{m})$ is a linear operator $\mathbf{I}: L^2(\mathbf{m}) \to L^2(\mathbf{m})$ such that for every function $f \in L^2(\mathbf{m})$,
\[
\mathbf{I}(f) = f.
\]
Furthermore, we define set-valued mapping
\begin{equation*}
	\mathbf{sign^*}(t)=
	\left\{
	\begin{aligned}
		1~~~~&t>0,\\
		-1~~~~&t<0,\\
		[-1,1]~~&t=0,
	\end{aligned}
	\right.
\end{equation*}
and signum function
\begin{equation*}
	\mathbf{sign}(t)=
	\left\{
	\begin{aligned}
		1~~~~&t\geq0,\\
		-1~~~~&t<0.
	\end{aligned}
	\right.
\end{equation*}

\begin{definition}[Covering numbers, Definition 2.1.5 of \cite{Vaart2023}]
	The \emph{covering number} $N(\varepsilon, \mathcal{F}, \|\cdot\|)$
	is the minimal number of balls $\{ g: \| g - f \| < \varepsilon \}$
	of radius $\varepsilon$ needed to cover the set $\mathcal{F}$.
	The centers of the balls need not belong to $\mathcal{F}$,
	but they should have finite norms.
	The \emph{entropy} (without bracketing) is the logarithm of the covering number.
\end{definition}
\begin{definition}[Bracketing numbers, Definition 2.1.6 of \cite{Vaart2023}]
	Given two functions $l$ and $u$, the \emph{bracket} $[l,u]$ is the set of all functions
	$f$ with $l \le f \le u$.
	An $\varepsilon$-bracket is a bracket $[l,u]$ with $\|u-l\| < \varepsilon$.
	The \emph{bracketing number} $N_{[\ ]}(\varepsilon, \mathcal{F}, \|\cdot\|)$
	is the minimum number of $\varepsilon$-brackets needed to cover $\mathcal{F}$.
	The \emph{entropy with bracketing} is the logarithm of the bracketing number.
	In the definition of the bracketing number, the upper and lower bounds $u$ and $l$
	of the brackets need not belong to $\mathcal{F}$ themselves but are assumed
	to have finite norms.
\end{definition}
\begin{definition}[Generalized normals, Definition 1.1 of \cite{mordukhovich2024second}]
	Let $\Omega$ be a nonempty subset of $X$. Given $x \in \Omega$ and $\varepsilon \geq 0$, define the set of $\varepsilon$-normals to $\Omega$ at $x$ by
	$$
	\widehat{N}_{\varepsilon}(x ; \Omega):=\left\{x^* \in X^* \left\lvert\, \limsup _{u \xrightarrow{\Omega} x} \frac{\left\langle x^*, u-x\right\rangle}{\|u-x\|} \leq \varepsilon\right.\right\} .
	$$
	When $\varepsilon=0$, elements of (1.2) are called Fréchet normals and their collection, denoted by $\widehat{N}(x ; \Omega)$, is the prenormal cone to $\Omega$ at $x$. If $x \notin \Omega$, we put $\widehat{N}_{\varepsilon}(x ; \Omega):=\emptyset$ for all $\varepsilon \geq 0$.
\end{definition}
\begin{definition}[Sequential Normal Compactness, Definition 1.20 of \cite{mordukhovich2006variational}]
A set $\Omega \subset X$ is Sequentially Normally Compact (SNC) at $\bar{x} \in \Omega$ if for any sequence $\left(\varepsilon_k, x_k, x_k^*\right) \in[0, \infty) \times \Omega \times X^*$ satisfying
	$$
	\varepsilon_k \downarrow 0, \quad x_k \rightarrow \bar{x}, \quad x_k^* \in \widehat{N}_{\varepsilon_k}\left(x_k ; \Omega\right), \quad \text { and } \quad x_k^* \xrightarrow{w^*} 0
	$$
	one has $\left\|x_k^*\right\| \rightarrow 0$ as $k \rightarrow \infty$.
\end{definition}
\begin{definition}[Sequential Normal Epi-Compactness of functions, Definition 1.116 of \cite{mordukhovich2006variational}]
 Let $\varphi: X \rightarrow \bar{\mathbb{R}}$ be finite at $\bar{x}$. We say that $\varphi$ is Sequentially Normally Epi-Compact (SNEC) at $\bar{x}$ if its epigraph is sequentially normally compact $a t(\bar{x}, \varphi(\bar{x}))$.
\end{definition}
\begin{definition}[Subderivatives, Definition 8.1 of \cite{Rockafellar1998}]
 For a function $f: \mathbb{R}^n \rightarrow \overline{\mathbb{R}}$ and a point $\bar{x}$ with $f(\bar{x})$ finite, the subderivative function $\mathbf{d} f(\bar{x}): \mathbb{R}^n \rightarrow \overline{\mathbb{R}}$ is defined by
	$$
	\mathbf{d}f(\bar{x})(\bar{w}):=\liminf _{\substack{\tau \downarrow 0 \\ w \rightarrow \bar{w}}} \frac{f(\bar{x}+\tau w)-f(\bar{x})}{\tau}
	$$
\end{definition}
\begin{definition}[Subdifferentials of extended-real-valued functions, Definition 1.32 of \cite{mordukhovich2024second}]\label{subgrad}
	 Let $\varphi: X \rightarrow \overline{\mathbb{R}}$ be an extended-real-valued function on a Banach space $X$.
	
	(i) Given $\varepsilon \geq 0$ and $x \in \operatorname{dom} \varphi$, the set
	$$
	\widehat{\partial}_{\varepsilon} \varphi(x):=\left\{x^* \in X^* \left\lvert\, \liminf _{u \rightarrow x} \frac{\varphi(u)-\varphi(x)-\left\langle x^*, u-x\right\rangle}{\|u-x\|} \geq-\varepsilon\right.\right\}
	$$
	is the $\varepsilon$-subdifferential of $\varphi$ at $x$. The set $\widehat{\partial}_0 \varphi(x)$ is denoted by
	$$
	\widehat{\partial} \varphi(x):=\left\{x^* \in X^* \left\lvert\, \liminf _{u \rightarrow x} \frac{\varphi(u)-\varphi(x)-\left\langle x^*, u-x\right\rangle}{\|u-x\|} \geq 0\right.\right\}
	$$
	and is called the presubdifferential or the regular subdifferential of $\varphi$ at this point. We put $\widehat{\partial}_{\varepsilon} \varphi(x):=\emptyset$ for all $\varepsilon \geq 0$ if $x \notin \operatorname{dom} \varphi$.
	
	(ii) Define the (basic, limiting)
	subdifferential of $\varphi$ at $\bar{x} \in \operatorname{dom} \varphi$ by
	\begin{equation*}
		\partial\varphi(\bar{x}) = \underset{\substack{x \xrightarrow{\varphi} \bar{x} \\ \varepsilon \downarrow 0}}{\operatorname{Lim\,sup}} \hat{\partial}_{\varepsilon}\varphi(x).
	\end{equation*}
	(iii) The singular subdifferential of $\varphi$ at $\bar{x} \in \operatorname{dom} \varphi$ is defined by
	\begin{equation*}
		\partial^{\infty}\varphi(\bar{x}) := \underset{\substack{x \xrightarrow{\varphi} \bar{x} \\ \varepsilon, \lambda \downarrow 0}}{\operatorname{Lim\,sup}} \lambda\hat{\partial}_{\varepsilon}\varphi(x).
	\end{equation*}
	We put $\partial\varphi(\bar{x}) := \emptyset$ and $\partial^{\infty}\varphi(\bar{x}) := \emptyset$ for $\bar{x} \notin \operatorname{dom} \varphi$.
\end{definition}
\begin{definition}[Clarke subdifferential, Definition 1 of \cite{bolte2007clarke}]
	The Clarke subdifferential $\partial^{C} f(x)$ of $f$ at $x$ is the set
	$$
	\partial^{C} f(x)= \begin{cases}\overline{\mathbf{conv}}\left\{\partial f(x)+\partial^{\infty} f(x)\right\} & \text { if } x \in \operatorname{dom} f, \\ \emptyset & \text { if } x \notin \operatorname{dom} f .\end{cases}
	$$
\end{definition}

\begin{definition}[Constructions of second-order subdifferentials, Definition 1.46 of \cite{mordukhovich2024second}]
	Let $\varphi: X \to \overline{\mathbb{R}}$ be an extended-real-valued function on a Banach space $X$, let $\bar{x} \in \mathrm{dom} \varphi$, and let $\bar{v} \in \partial \varphi(\bar{x})$ be a first-order subgradient from Definition \ref{subgrad}. Define:
	\begin{enumerate}
		\item[(i)] The mapping $\partial^2_N \varphi(\bar{x}, \bar{v}): X^{**} \rightrightarrows X^*$ with the values
		\begin{equation*}
			\partial^2_N \varphi(\bar{x}, \bar{v})(u) := (D_N^* \partial \varphi)(\bar{x}, \bar{v})(u), \quad u \in X^{**},
		\end{equation*}
		is the normal second-order subdifferential of $\varphi$ at $\bar{x}$ relative to $\bar{v}$.
		\item[(ii)] The mapping $\partial^2_M \varphi(\bar{x}, \bar{v}): X^{**} \rightrightarrows X^*$ with the values
		\begin{equation*}
			\partial^2_M \varphi(\bar{x}, \bar{v})(u) := (D_M^* \partial \varphi)(\bar{x}, \bar{v})(u), \quad u \in X^{**},
		\end{equation*}
		is the mixed second-order subdifferential of $\varphi$ at $\bar{x}$ relative to $\bar{v}$.
		
		\item[(iii)] The mapping $\check{\partial}^2 \varphi(\bar{x}, \bar{v}): X^{**} \rightrightarrows X^*$ with the values
		\begin{equation*}
			\check{\partial}^2 \varphi(\bar{x}, \bar{v})(u) := (\hat{D}^* \partial \varphi)(\bar{x}, \bar{v})(u), \quad u \in X^{**},
		\end{equation*}
		is the combined second-order subdifferential of $\varphi$ at $\bar{x}$ relative to $\bar{v}$.
	\end{enumerate}
	For the definition of co-derivative $D^*_N$, see \cite{mordukhovich2024second}.
\end{definition}
\begin{definition}[Lower closure, Page 14 of \cite{Rockafellar1998}]
	The function is lower semi-continuous and is the greatest of all the lower semi-continuous functions $g$ such that $g\leq f$. It is called the lower closure of $f$, denoted by ${\rm cl}f$.
\end{definition}

\begin{definition}[Lower and upper epi-limits, Definition 7.1 of \cite{Rockafellar1998}]
	For any sequence $\{f^v\}_{v\in\mathbb{N}}$ of functions on $\mathbb{R}^n$, the lower epi-limit ${\rm e\text{-}liminf}_v f^v$ is the function having as its epigraph the outer limit of the sequence of sets ${\rm epi}f^v$:
	$${\rm epi}({\rm e\text{-}liminf}_v f^v)\doteq {\rm limsup}_v({\rm epi}f^v). $$
	The upper epi-limit ${\rm e\text{-}limsup}_v f^v$ is the function having as its epigraph the inner limit of the sets ${\rm epi}f^v$:
	$${\rm epi}({\rm e\text{-}limsup}_v f^v)\doteq {\rm liminf}_v({\rm epi}f^v). $$
	When these two functions coincide, the epi-limits function ${\rm e\text{-}lim}f^v$ is said to exist:
	$${\rm e\text{-}lim}f^v\doteq{\rm e\text{-}liminf}_v f^v={\rm e\text{-}limsup}_v f^v.$$
\end{definition}

\begin{definition}[O-minimal structure, Definition 6 of \cite{bolte2007clarke}]
	{
		An {o-minimal structure} is a sequence of Boolean algebras $\mathcal{O}_d$ of subsets of $\R^d$ such that for each $d\in \mathbb{N}$:
		\begin{enumerate}[(i)]
			\item if $A$ belongs to $\mathcal{O}_d$, then $A\times \R$ and $\R\times A$ belong to $\mathcal{O}_{d+1}$;
			\item if $\pi\colon\R^{d}\times \R\to\R^d$ denotes the coordinate projection onto $\R^d$, then for any $A$ in $\mathcal{O}_{d+1}$ the set $\pi(A)$ belongs to  $\mathcal{O}_{d}$;
			\item  $\mathcal{O}_{d}$ contains all sets of the form $\{x\in \R^d: p(x)=0\}$, where $p$ is  a polynomial on $\R^d$;
			\item the elements of $\mathcal{O}_{1}$ are exactly the finite unions of intervals (possibly infinite) and points.
		\end{enumerate}
		The sets $A$ belonging to $\mathcal{O}_{d}$, for some $d\in \mathbb{N}$, are called { definable in the o-minimal structure}.}
\end{definition}

\begin{definition}[Lyapunov condition, Assumption B of \cite{davis2020stochastic}]
Let $\mathcal{X}$ be a closed set and let $G: \mathcal{X} \rightrightarrows \mathbb{R}^d$ be a set-valued map. Then an arc $z: \mathbb{R}_{+} \rightarrow \mathbb{R}^d$ is called a trajectory of $G$ if it satisfies the differential inclusion
	$\dot{z}(t) \in G(z(t)) \text { for a.e. } t \geq 0$.
 there exists a continuous function $\varphi: \mathbb{R}^d \rightarrow \mathbb{R}$, which is bounded from below, and such that the following two properties hold.
	\begin{itemize}
		\item (Weak Sard) For a dense set of values $r \in \mathbb{R}$, the intersection $\varphi^{-1}(r) \cap G^{-1}(0)$ is empty.
		\item (Descent) Whenever $z: \mathbb{R}_{+} \rightarrow \mathbb{R}^d$ is a trajectory of the differential inclusion and $0 \notin G(z(0))$, there exists a real $T>0$ satisfying
		$$
		\varphi(z(T))<\sup _{t \in[0, T]} \varphi(z(t)) \leq \varphi(z(0)) .
		$$
	\end{itemize}
\end{definition}
\begin{definition}[Chain rule, Definition 5.1 of \cite{davis2020stochastic}]
	Consider a locally Lipschitz function $f$ on $\mathbb{R}^d$. We will say that $f$ admits a chain rule if for any absolutely continuous curves $z: \mathbb{R}_{+} \rightarrow \mathbb{R}^d$, equality
	$$
	(f \circ z)^{\prime}(t)=\langle\partial^C f(z(t)), {z}^{\prime}(t)\rangle \text { holds for a.e. } t \geq 0,
	$$
\end{definition}
\begin{definition}[Smooth manifold, tangent space and normal space, Page 13, Page 51 and Page 138 of \cite{lee2003smooth}]
A set $M \subset \mathbb{R}^d$ is a $C^p$ smooth manifold if there is an integer $r \in \mathbb{N}$ such that around any point $x \in M$, there is a neighborhood $U$ and a $C^p$-smooth map $F: U \rightarrow \mathbb{R}^{d-r}$ with $\nabla F(x)$ of full rank and satisfying $M \cap U=\{y \in U: F(y)=0\}$. If this is the case, the tangent and normal spaces to $M$ at $x$ are defined to be $T_M(x):=\operatorname{Null}(\nabla F(x))$ and $N_M(x):=\left(T_M(x)\right)^{\perp}$, respectively.
\end{definition}
\begin{definition}[Whitney stratification, Definition 5.6 of \cite{davis2020stochastic}]
A Whitney $C^p$-stratification $\mathcal{A}$ of a set $Q \subset \mathbb{R}^d$ is a partition of $Q$ into finitely many nonempty $C^p$ manifolds, called strata, satisfying the following compatibility conditions.
	\begin{itemize}
		\item  Frontier condition: For any two strata $L$ and $M$, the implication
		$$
		L \cap \operatorname{cl} M \neq \emptyset \quad \Longrightarrow \quad L \subset \operatorname{cl} M \text { holds. }
		$$
		\item  Whitney condition: For any sequence of points $z_k$ in a stratum $M$ converging to a point $\bar{z}$ in a stratum $L$, if the corresponding normal vectors $v_k \in N_M\left(z_k\right)$ converge to a vector $v$, then the inclusion $v \in N_L(\bar{z})$ holds.
	\end{itemize}
		A function $f: \mathbb{R}^d \rightarrow \mathbb{R}$ is Whitney $C^p$-stratifiable if its graph admits a Whitney $C^p$-stratification.
\end{definition}
\begin{definition}
 [H{\"o}lder formulation class, Page 7 of \cite{schmidt2020nonparametric}] Let $\gamma$ and $B$ be two positive constants and $\lfloor \gamma \rfloor$ denote the largest integer strictly less than $\gamma$. We call a function $h: \mathbb{T}\subset \mathbb{R}^q\rightarrow\mathbb{R}$ a  $(\gamma,B)$-H{\"o}lder smooth function if it satisfies
	\begin{equation*}
		\sup_{z\in \mathbb{T}}    \Big|\frac{\partial^{|\boldsymbol{\alpha}|}h}{\partial^{\alpha_1}z_1\ldots\partial^{\alpha_q}z_q}(z)\Big|\le B,~\text{for all}~ \boldsymbol{\alpha}=(\alpha_1,\ldots,\alpha_q)^\top\in\mathbb{N}^q ~\text{and}~|\boldsymbol{\alpha}|=\sum_{i=1}^{q}\alpha_i\le \lfloor \gamma \rfloor,
	\end{equation*}
	and
	\begin{equation*}
		\mathop{\sup}_{z,z^*\in\mathbb{T}} \Big|\frac{\partial^{|\boldsymbol{\alpha}|}h}{\partial^{\alpha_1}z_1\ldots\partial^{\alpha_q}z_q}(z )-\frac{\partial^{|\boldsymbol{\alpha}|}h}{\partial^{\alpha_1}z_1\ldots\partial^{\alpha_q}z_q}(z^*)
		\Big|\le B\|z-z^*\|_{2}^{\gamma-\lfloor \gamma \rfloor},~\text{for all}~|\boldsymbol{\alpha}|=\lfloor \gamma \rfloor.
	\end{equation*}
	Denote the class of all such $(\gamma,B)$-H{\"o}lder smooth functions as $\mathcal{H}_{q}^{\gamma}(\mathbb{T},B)$. Let 
	$J\in \mathbb{N}$, $\boldsymbol{\gamma}=(\gamma_1,\ldots,\gamma_{J})^\top\in\mathbb{R}_{+}^{J}$,  $\boldsymbol{d}=(q,d_1,\ldots,d_{J})^\top\in\mathbb{N}^{J+1}$ and $\boldsymbol{\bar{d}}=(\bar{d}_1,\ldots,\bar{d}_{J})^\top\in\mathbb{N}^{J}$ with $\bar{d}_1\le q$ and $\bar{d}_k\le d_{k-1}, k=2,\ldots,J$. We further define a composite function class:
	\begin{equation}\label{def: calH}
		\begin{aligned}
			\mathcal{H}(J,\boldsymbol{\gamma},{\boldsymbol{d}},\boldsymbol{\bar{d}},B)=\Big\{h=&h_{J}\circ\ldots\circ h_1:\mathbb{T}\rightarrow\mathbb{R}~|~ \   h_k=(h_{k1},\ldots,h_{k d_k})^\top~\text{and}~\\& h_{kj}\in \mathcal{H}_{\bar{d}_k}^{\gamma_k}([a_k,b_k]^{\bar{d}_k},B) ~\text{for some}~|a_k|,|b_k|\le B \Big\}.
		\end{aligned}
	\end{equation}
	 We call   $\boldsymbol{\bar{d}}$ the \textit{intrinsic dimension} of the function $h$ in $\mathcal{H}(J,\boldsymbol{\gamma},{\boldsymbol{d}},\boldsymbol{\bar{d}},B)$.
\end{definition}
\begin{definition}
	[Dual operator, Theorem 5.11-1 of \cite{ciarlet2025linear}] Let $X$ and $Y$ be two normed vector spaces over the same field $\mathbb{K}$. Given any operator $A \in \mathcal{L}(X ; Y)$, there exists one and only one operator $A^* \in \mathcal{L}\left(Y^* ; X^*\right)$, called the dual operator of $A$, or simply the dual of $A$, such that
	$$
	A^* y^*(x)=y^*(A x) \quad \text { for all } x \in X \text { and all } y^* \in Y^* .
	$$
Besides,
	$
	\left\|A^*\right\|_{\mathcal{L}\left(Y^* ; X^*\right)}=\|A\|_{\mathcal{L}(X ; Y)} .
	$
\end{definition}
\textbf{Notations of Deep Neural Networks (DNN).} Let $L \ge 2$ be an integer representing the number of layers, and let $\boldsymbol{q}=(q_0,q_1,\ldots,q_L)^\top\in\mathbb{N}^{L+1}$ define the number of neurons in each layer. An $L$-layer neural network is a function $g:\mathbb{R}^{q_0}\rightarrow \mathbb{R}^{q_{L}}$ that maps a $q_0$-dimensional input to a $q_L$-dimensional output. It is defined by the following composition of functions:
\begin{equation}\label{def: DNN1}
	\begin{aligned}
		& m_0(z) = z, \\
		&m_1(z)=\sigma_{1}(W_1m_0(z)+b_1),\\
		&\ldots,\\
		&m_{L-1}(z) = \sigma_{L-1}(W_{L-1}m_{L-2}(z)+b_{L-1}), \\
		& g(z)=W_L m_{L-1}(z)+b_L,
	\end{aligned}
\end{equation}
where for each layer $k=1,\ldots,L$, $W_k$ is a $q_{k}\times q_{k-1}$ weight matrix and $b_k$ is a $q_k$-dimensional bias vector. The term $m_k$ for $1\le k\le L-1$ represents the output of the $k$-th \textit{hidden layer}, and $L$ is the \textit{depth} of the network.
The functions $\sigma_k: \mathbb{R}^{q_{k}} \to \mathbb{R}^{q_{k}}$ for $k=1,\ldots,L-1$ are activation functions that operate element-wise on their input vectors. That is, for a vector $v=(v_1,\ldots,v_{q_k})^\top$, $\sigma_k(v) = (\sigma(v_1),\ldots,\sigma(v_{q_k}))^\top$, where $\sigma: \mathbb{R}\to\mathbb{R}$ is a scalar activation function (e.g., ReLU, Sigmoid). Note that the final layer (layer $L$) has no activation function, a common configuration for regression tasks.
To simplify the notation, we can absorb the bias vector $b_k$ into the weight matrix $W_k$. This is achieved by defining an augmented weight matrix $\tilde{W}_k=(W_k, b_k)\in\mathbb{R}^{q_k\times (q_{k-1}+1)}$ and appending a 1 to the input of each layer. For instance, the network's input $z$ is augmented to $\tilde{z}=(z^\top,1)^\top$.
This requires a corresponding modification of the activation functions. For each hidden layer $k=1,\ldots,L-1$, we define an operator $\phi_k$ that first applies the activation $\sigma_k$ and then appends a 1 to the resulting vector:
$$
\phi_k(v) = (\sigma_k(v)^\top, 1)^\top.
$$
With these definitions, the neural network in \eqref{def: DNN1} can be expressed more compactly as a composition of matrix-vector products and activation operators:
\begin{equation}\label{def: DNN2}
	g(z)=\tilde{W}_{L} \circ \phi_{L-1} \circ \tilde{W}_{L-1} \circ \cdots \circ \phi_1 \circ \tilde{W}_1(\tilde{z}).
\end{equation}


\begin{figure}[h!]
\centering
\begin{tikzpicture}[x=2.2cm, y=1.2cm, >=stealth]

\definecolor{inputcolor}{RGB}{60,120,216}
\definecolor{hiddencolor}{RGB}{230,90,90}
\definecolor{outputcolor}{RGB}{50,90,200}

\foreach \i in {1,2,3,4}
{
    \node[circle, draw, fill=inputcolor, minimum size=10pt]
        (I\i) at (0, -0.5-\i) {};
    \node[left=2pt of I\i] {$z_{\i}$};
}
\node at (0,-6.3) {input-layer};

\foreach \i in {1,2,3,4,5}
{
    \node[circle, draw, fill=hiddencolor, minimum size=10pt]
        (H1\i) at (2, -\i) {};
}
\node at (2,-6.3) {hidden-layer 1};

\foreach \i in {1,2,3,4,5}
{
    \node[circle, draw, fill=hiddencolor, minimum size=10pt]
        (H2\i) at (4, -\i) {};
}
\node at (4,-6.3) {hidden-layer 2};

\node[circle, draw, fill=outputcolor, minimum size=10pt]
    (O) at (6, -3) {};
\node[right=4pt of O] {$m(z)$};
\node at (6,-6.3) {output-layer};


\foreach \i in {1,2,3,4}
    \foreach \j in {1,...,5}
        \draw[->] (I\i) -- (H1\j);

\foreach \i in {1,...,5}
    \foreach \j in {1,...,5}
        \draw[->] (H1\i) -- (H2\j);

\foreach \i in {1,...,5}
    \draw[->, thick] (H2\i) -- (O);

\end{tikzpicture}
\caption{\label{fig: network} A 3-layer neural network with four input variables and one output.}
\end{figure}
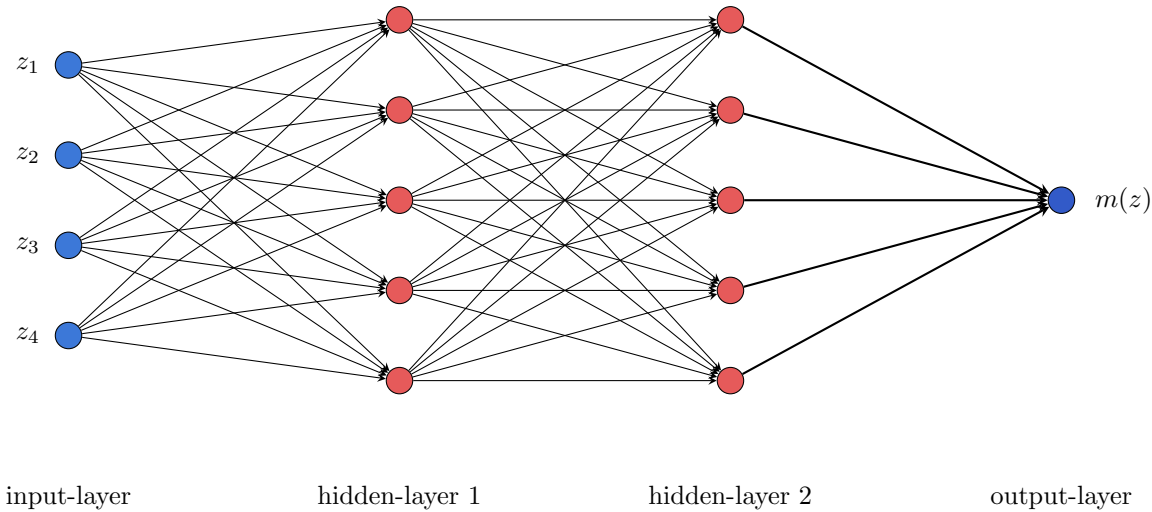

Note that the total number of parameters in \eqref{def: DNN2} is $\sum_{k=1}^{L}q_{k}(q_{k-1}+1)$, which can be very large and may lead to overfitting.
For $s\in \mathbb{N}$, $L\ge 2$, $A> 0$ and $\boldsymbol{q}=(q_0,q_1,\ldots,q_L)^\top$, we consider a sparsely connected neural network class
\begin{equation}\label{def: sparse NN}
\begin{aligned}
    \mathcal{M}(s,L,\boldsymbol{q},A)=\Big\{ g(z)=&W_{L}\phi_{L-1}\circ\cdots\circ W_2\phi_1(W_1 \tilde{z})~|~ W_k\in\mathbb{R}^{q_k\times(q_{k-1}+1)},~  \|W_k\|_{\infty}\le 1~\text{for}~ \\
    & k =1,\ldots,L,~ \sum_{k=1}^{L}\|W_k\|_{0}\le s ~\text{and}~ \|g\|_{\infty}\le A    \Big\},
\end{aligned}
\end{equation}
where $\|\cdot\|_{\infty}$ is the sup-norm of a matrix or function and $\|\cdot\|_{0}$ is the number of non-zero elements of a matrix.

\section{Statistical Perspective}\label{sec3}

To solve \eqref{opt0}, we parametrize the nonlinear term $g$ by DNN and rewrite the problem \eqref{opt0} as
\begin{equation}\label{opt1}
	\min_{\beta\in\mathbb{R}_C^d, g\in\mathcal{M}(s, L, \boldsymbol{q}, C)} \frac{1}{N} \sum_{i=1}^{N} \left|Y_i - \beta^\top X_i - g(Z_i)\right|+\lambda_N \mathcal{J}_{N,M}(\beta,g)
\end{equation}
with parameters of DNN $\mathbf{W}:=(W_1,\dots,W_L)$.  Due to the DNN structure of $\mathcal{M}$, \eqref{opt1} can be rewritten into the following finite-dimension form:
\begin{equation}\label{opt2}
\begin{aligned}
	\min_{\beta, \mathbf{W}}~~& \frac{1}{N} \sum_{i=1}^{N} \left|Y_i - \beta^\top X_i - g(\mathbf{W};Z_i)\right|+\lambda_N \mathcal{J}_{N,M}(\beta,g(\mathbf{W})),\\
\mathrm{s.t.}~~~~~~& \|W_k\|_{\infty}\le 1, ~\text{for}~k =1,\ldots,L,\\
&\sum_{k=1}^{L}\|W_k\|_{0}\le s
\end{aligned}	
\end{equation}
for given $L, s$.  $\mathcal{J}_{N,M}(\beta,g(\mathbf{W}))$ (or equivalently denoted as $\mathcal{J}_{N,M}(\beta,\mathbf{W})$) can be any bounded regulation term. For example $\mathcal{J}_{N,M}(\beta,g(\mathbf{W}))=|\partial_{Z}g(\mathbf{W};Z)|\wedge M$.

\begin{remark}
	The norm of the Jacobian of $g$ with respect to the input $Z$ is used as a penalty term in DNN training to prevent overfitting and improve robustness against input data corruption \cite{hoffman2019robust}. Furthermore, this Jacobian term can be efficiently computed using the standard backpropagation algorithm \cite{AMARI1993185}.
\end{remark}

For some $J\in \mathbb{N}$, $\boldsymbol{\gamma}=(\gamma_1,\ldots,\gamma_{J})\in\mathbb{R}_{+}^{J}$,  $\boldsymbol{d}=(q,d_1,\ldots,d_{J})^\top\in\mathbb{N}^{J+1}$ and $\boldsymbol{\bar{d}}=(\bar{d}_1,\ldots,\bar{d}_{J})^\top\in\mathbb{N}^{J}$ with $\bar{d}_1\le q$ and $\bar{d}_k\le d_{k-1}, k=2,\ldots,J ,$  we define the \textit{effective smoothness}  $\bar{\gamma}_{k}= \gamma_{k}\prod_{i=k+1}^{J}(\gamma_i\wedge 1)$  of a function $h$ in $\mathcal{H}(J,\boldsymbol{\gamma},{\boldsymbol{d}},\boldsymbol{\bar{d}},B),$  and write
\begin{equation*}
	\zeta=\mathop{\min}_{k\in\{1,\ldots,J\}} \frac{\bar{\gamma}_{k}}{2\bar{\gamma}_{k}+\bar{d}_k}~\text{and}~r_N=N^{-\zeta}.
\end{equation*}
For the covariate $X=(X_{(1)},\ldots,X_{(d)})^\top$, we define 
\begin{equation}\label{def: projection}
	\varphi_{k}^{*}=\mathop{\arg\min}_{\varphi\in L^{2}(P_{Z})}\mathbb{E} [f_\varepsilon(0|V)\{X_{(k)}-\varphi(Z)\}^2], k=1,\ldots,d,
\end{equation}
where $L^{2}(P_{Z})=\{\varphi~|~ \mathbb{E}\varphi^2(Z) < \infty \}.$ And denote  
$\boldsymbol{\varphi}^*(Z)=(\varphi_1^*(Z),\ldots,\varphi_d^*(Z))^\top,$   $\Sigma_1=\mathbb{E}[\{X-\boldsymbol{\varphi}^*(Z)\}\{X-\boldsymbol{\varphi}^*(Z)\}^\top]$ and  $\Sigma_2=\mathbb{E}[f_\varepsilon(0|X,Z)\{X-\boldsymbol{\varphi}^*(Z)\}\{X-\boldsymbol{\varphi}^*(Z)\}^\top].$


We consider the joint probability space $(\Omega, \mathcal{F}, \widehat{P})$ for the random variables $(X, Z, \varepsilon)$.
Here, the sample space $\Omega = \Omega_X \times \Omega_Z \times \Omega_\varepsilon$ is the product of their sample spaces,
$\mathcal{F}$ is the corresponding product $\sigma$-algebra, and $\widehat{P}$ is their joint probability measure.  Furthermore, we define the probability space $(\Omega^\infty:=\prod_{i=1}^{\infty}\Omega^i, \mathcal{F}^\infty:=\sigma(\prod_{i=1}^{\infty}\mathcal{F}^i), P)$
for the sequence of samples $((X_i,Z_i,\varepsilon_i))_{i=1}^\infty$,
where $\Omega^\infty$ denotes the product space and $\mathcal{F}^\infty$ is the corresponding product $\sigma$-algebra. Furthermore, we assume $P(\varepsilon\leq0)=\frac{1}{2}$. For simplicity in the subsequent analysis, we let $\mathbb{E}[\cdot]=\mathbb{E}_U[\cdot]$.

Let $\mathbb{R}_{C}^{d}=\{\beta\in\mathbb{R}^{d}~|~\|\beta\|_\infty<C \}$, we define
\begin{equation}\label{def estimator}
	(\hat{\beta}_N,\hat{g}_N)\in \mathop{\arg\min}_{\beta\in\mathbb{R}_C^d,g\in\mathcal{M}_C^N} L_N(\theta)+\lambda_N \mathcal{J}_{N,M}(\beta,g)
\end{equation}
with $L_N(\theta):=\frac{1}{N} \sum_{i=1}^{N} \left|Y_i - \beta^\top X_i - g(Z_i)\right|$.

\begin{assumption}
We introduce the following assumptions.
\begin{enumerate}[{({A}1)}]
	\item The true vector parameter $\beta_0$ belongs to a compact subset $\mathbb{R}_C^d:=\{\beta\in\mathbb{R}^d\mid \Vert \beta \Vert_\infty<C\}$ and the true nonparametric function $g_0$ satisfies $\Vert g_0 \Vert_\infty<C$ and belongs to $\mathcal{H}=\mathcal{H}(J,\boldsymbol{\gamma},{\boldsymbol{d}},\boldsymbol{\bar{d}},B)$.  \label{assump: param space} 
	\item There exists a constant $A_0>0$ s.t. $\sigma_{\min}(\Sigma_2)>A_0$ and $\sigma_{\min}(\mathbb{E}[(X-\mathbb{E}[X|Z])(X-\mathbb{E}[X|Z])^\top])>A_0$.
	\item The covariates   $V=(X,Z)$ take values in a compact subset of $\mathbb{R}^{d+l}$ that, without loss of generality, will be assumed to be $[0,1]^{d+l}.$ In addition, the probability density function (PDF) of $Z$ is bounded away from zero and from infinity. \label{assump: covariates}
	\item $L =O(\log N),$ $s=O(N r_N^2 \log N)$, $\lambda_N=o(1)$ and
	\begin{equation*}
		Nr_N^2\lesssim \underset{k=1,\ldots,L}{\min}\{q_k\}\le\!\underset{k=1,\ldots,L}{\max}\{q_k\} \lesssim N.
	\end{equation*}\label{assump: neural struc}
	\item The conditional PDF $f_\varepsilon(\cdot|v)$   of the random error $\varepsilon$ given the covariate $V=v$, has continuous derivative $f^{'}_\varepsilon(\cdot|v),$ and there exist positive constants $b_0$ and $c_0$ 
	such that $1/c_0<f_\varepsilon(t|v) < c_0$ and $|f^{'}_\varepsilon(t|v)|<d_0$ for all $|t|\le b_0, v\in[0,1]^{d+l}.$ Furthermore, we assume $\mathbb{E}[|\varepsilon||V=v]<\infty$ for any $v\in [0,1]^{d+l}$. \label{assump: error pdf}
	\item  For any $k\in\{1,\dots,J\}$, $\bar{\gamma}_{{k}}>\bar{d}_{{k}}/2$, and $\mathbb{E}[\Vert X\Vert^2]< \infty$. \label{assump: AsympNormal}
	\item   In addition, there exists $B_f>0$ such that $f(t|v)\leq B_f$ for all $t\in\mathbb{R}$ and $v\in[0,1]^{d+l}$. \label{assump:rate_lambda}
	\item $\mathcal{J}_{N,M}$ is separable i.e. $\mathcal{J}_{N,M}(\beta,g)=\mathcal{J}_{N,1}(\beta)+\mathcal{J}_{N,2}(g)$; $\mathcal{J}_{N,M}$ is lower semi continuous (l.s.c.) and $S N E C$ on $\mathbb{R}_C^d\times \mathcal{M}_C^N$, and let the qualification condition
	$$
	\left[\left(0, \mathbf{v}^*\right) \in \partial^{\infty}_{(\xi,h)}\mathcal{J}_{N,M}(\beta,g)\right] \Longrightarrow \mathbf{v}^*=0
	$$
	with ${\xi}={\beta}-\beta_0,$ ${h}(Z)={g}(Z)-g_0(Z)+(\beta-\beta_0)^\top {\varphi}^*(Z)$
	for any $(\xi,h)\in\mathbb{R}_C^d\times \mathcal{M}_C^N$; $\lambda_N\left(\| \partial_{g}\mathcal{J}_{N,2}\|_*+\|\partial_{\beta}\mathcal{J}_{N,1}\|\right)=o_p\left(\frac{1}{\sqrt{N}}\right)$. Here, $\|\cdot\|_*$ denotes the operator norm in $L_2(\mathbf{m})$ space with Lebesgue measure $\mathbf{m}$. \label{assump: regular}
\end{enumerate}

\end{assumption}

\begin{theorem}\label{thm1}
	Suppose Assumptions (A\ref{assump: param space})-(A\ref{assump: error pdf}) hold. Then the estimators $\hat{\beta}_N$ and $\hat{g}_N$ from optimization problem \eqref{def estimator} exhibit the following rates of convergence:
	\begin{equation*}
		\begin{aligned}
			\|\hat{\beta}_N-\beta_0\|_\infty &= O_p(r_N\log^2 N+\lambda_N), \\
			\|\hat{g}_N-g_0\|_{L^2(P)} &= O_p(r_N\log^2 N+\lambda_N).
		\end{aligned}
	\end{equation*}
\end{theorem}
\begin{proof}
	Let  $\hat{\theta}_N =(\hat{\beta}_N,\hat{g}_N)$, $\theta_0=(\beta_0,g_0)$ and $d(\theta_1,\theta_2)=[\mathbb{E}\{X^\top\beta_1+g_1(Z)-X^\top\beta_2-g_2(Z)\}^2]^{1/2}$, for any $\theta_1=(\beta_1,g_1)$ and $\theta_2=(\beta_2,g_2)$.  We first show that
	\begin{equation}\label{consi}
		d(\hat{\theta}_N,\theta_0)\overset{p}{\rightarrow} 0,\  \text{as}\ N \to \infty.
	\end{equation}
We first show that
\begin{equation}\label{pfthm1: uniform consistemcy}
	\mathop{\sup}_{\theta \in \mathbb{R}^{d}_{C}\times\mathcal{M}_C^N}|L_{N}(\theta)+\lambda_N\mathcal{J}_{N,M}(\theta)-L_{0}(\theta)|\xrightarrow{p}0,~\text{as}~N\rightarrow \infty.
\end{equation}
with $L_0(\theta):=\mathbb{E}\left[\left|Y-\beta^\top X -g(Z)\right|\right]$.
Based on the assumption of $\lambda_N$ and the definition of $\mathcal{J}_{N,M}(\theta)$, it suffices to show
\begin{equation}\label{GC}
	\mathop{\sup}_{\theta \in \mathbb{R}^{d}_{C}\times\mathcal{M}_C^N}|L_{N}(\theta)-L_{0}(\theta)|\xrightarrow{p}0,~\text{as}~N\rightarrow \infty.
\end{equation}
 Denote $\mathcal{F}_N:=\{f(x,y,z):=|y - \beta^\top x - g(z)|,\ \forall (x,y,z)\in \mathbb{R}\times[0,1]^{d+l}~|~(\beta,g)\in\mathbb{R}^{d}_C\times\mathcal{M}_C^N\}$. Notably, $F(x,y,z):=|y|+2C$ is an envelope function of $\mathcal{F}_N$ with $\mathbb{E} F <\infty$.  Based on Theorem 2.4.6 of \cite{Vaart2023}, it holds that
\begin{equation}\label{GC_thm}
	\begin{aligned}
		&\mathbb{E}^*\|\mathbb{P}_N-P\|_{\mathcal{F}_N}\\
	    \overset{(\rm I)}{\leq}& \mathbb{E}^*_{U,e}\left\Vert\frac{1}{N}\sum_{i=1}^{N}e_i f(U_i)\right\Vert_{\mathcal{F}_N}\\
		\overset{(\rm II)}{=}& 2\mathbb{E}\mathbb{E}_e \left\Vert\frac{1}{N}\sum_{i=1}^{N}e_if(U_i)\right\Vert_{\mathcal{F}_N}\\
		\overset{(\rm III)}{\leq}& 2\mathbb{E}\mathbb{E}_e \left\Vert\frac{1}{N}\sum_{i=1}^{N}e_if(U_i)\right\Vert_{\mathcal{F}_N\wedge q}+\mathbb{E} F\mathbf{1}\{F>q\}\\
		 \overset{(\rm IV)}{\leq}& 2\mathbb{E}\left\{\sqrt{1+\log N(\epsilon, \mathcal{F}_N\wedge q, L_1(\mathbb{P}_N))} \sup_{f\in\mathcal{G}_N}\left\Vert \frac{1}{N} \sum_{i=1}^{N}e_i f(U_i)\right\Vert_{\psi_2\mid U}\right\}+\underbrace{2{\epsilon}+\epsilon_F}_{:=\hat{\epsilon}}\\
		 \overset{}{\leq}& 2\mathbb{E}\left\{\sqrt{1+\log N(\epsilon, \mathcal{F}_N\wedge q, L_1(\mathbb{P}_N))}\sqrt{\frac{6}{N}}q\right\}+\hat{\epsilon}
	\end{aligned}
\end{equation}
with $\mathcal{F}_N\wedge q=\{f\wedge q\mid f\in\mathcal{F}_N\}$.
Here, inequality ${(\rm I})$ holds due to Lemma 2.3.1 of \cite{Vaart2023}, where $e_1,\dots, e_N$ are i.i.d. Rademacher random variables.
The inequality $(\rm II)$ holds due to the measurability of $\left\Vert\frac{1}{N}\sum_{i=1}^{N}e_if(U_i)\right\Vert_{\mathcal{F}_N}$ and Fubini Theorem. Indeed, Schmidt-Hieber (2020) proves
\begin{equation}\label{lemma 5}
	\log N(\epsilon, \mathcal{M}_C^N, \Vert\cdot\Vert_\infty)\leq (s+1)\log\left(\frac{2H^2(L+1)}{\epsilon}\right)
\end{equation}
with $H:=\prod_{k=1}^{L}(q_k+1)$;
see Lemma 5 of \cite{10.1214/19-AOS1875} for details. Combining with the fact that $\beta\in\mathbb{R}_C^d$, we may conclude that there exists a countable dense subset $\{\tilde{f}_i\}_{i\geq 0}$ of $\mathcal{F}_N$. Thus, $\left\Vert\frac{1}{N}\sum_{i=1}^{N}e_if(U_i)\right\Vert_{\mathcal{F}_N}$ equals $\left\Vert\frac{1}{N}\sum_{i=1}^{N}e_if(U_i)\right\Vert_{\{\tilde{f}_i\}_{i\geq 0}}$, and is of course $P$-measurable.
{Therefore,  in this paper, we no need to distinguish the outer measure (expectation) and classic measure (expectation).}
   Furthermore, we denote $\mathcal{G}_N$ is an $\epsilon$-net in $L_1(\mathbb{P}_N)$ over $\mathcal{F}_N\wedge q$, while inequality ${(\rm IV)}$ holds for any $\epsilon_F>0$ by selecting sufficiently large $q$.

Due to the triangle inequality, $\forall f_1,f_2\in\mathcal{F}_N\wedge q$, we have
\begin{equation*}
	\Vert f_1-f_2\Vert_\infty\leq \sup_{\forall x,z\in[0,1]^{d+l}}|g_1(z)-g_2(z)+(\beta_1-\beta_2)^\top x |\leq \Vert g_1-g_2\Vert_\infty+\Vert \beta_1-\beta_2\Vert_\infty.
\end{equation*}
Hence, by \eqref{lemma 5}, it is easy to show that
\begin{equation}\label{covering num}
\begin{aligned}
		N(\epsilon, \mathcal{F}_N\wedge q, L_1(\mathbb{P}_N))\leq &N(\epsilon, \mathcal{F}_N\wedge q, \Vert\cdot\Vert_\infty) \\
		\leq & N\left(\frac{\epsilon}{2}, \mathcal{M}_C^N,\Vert\cdot\Vert_\infty\right)N\left(\frac{\epsilon}{2},\mathbb{R}_C^d,\Vert\cdot\Vert_\infty\right)\\
		{\leq}& K_\epsilon N\left(\frac{\epsilon}{2}, \mathcal{M}_C^N,\Vert\cdot\Vert_\infty\right),
\end{aligned}
\end{equation}
where the first inequality holds due to $\Vert f\Vert_{L^1(\mathbb{P}_N)}\leq \Vert f \Vert_\infty,$ for any $ f\in\mathcal{F}_N\wedge q$ and discrete measure $\mathbb{P}_N$ combining with the nature of $\epsilon$-net. Moreover, the second inequality holds due to Heine–Borel Theorem in which $ K_\epsilon=N\left(\frac{\epsilon}{2},\mathbb{R}_C^d,\Vert\cdot\Vert_\infty\right)$.
 Substituting \eqref{covering num} into \eqref{GC_thm}, it holds that
\begin{equation*}
	\begin{aligned}
				\mathbb{E}\|\mathbb{P}_N-P\|_{\mathcal{F}_N}\leq& 2\mathbb{E}\left\{\sqrt{1+\log\left(K_\epsilon N\left(\frac{\epsilon}{2}, \mathcal{M}_C^N,\Vert\cdot\Vert_\infty\right)\right)}\sqrt{\frac{6}{N}}q\right\}+\hat{\epsilon}	 \\
				{\leq}& 2\mathbb{E}\left\{\sqrt{1+\log(K_\epsilon)+(s+1)\log\left(\frac{4H^2(L+1)}{\epsilon}\right)}\sqrt{\frac{6}{N}}q\right\}+\hat{\epsilon}\quad (\text{By} {\eqref{lemma 5}}).
	\end{aligned}
\end{equation*}
Due to Assumption (A\ref{assump: neural struc}), the integrand is
\begin{equation*}
\begin{aligned}	\sqrt{1+\log(K_\epsilon)+(s+1)\log\left(\frac{4H^2(L+1)}{\epsilon}\right)}\sqrt{\frac{6}{N}}q=O(r_N\log^{\frac{3}{2}} N)=o(1).
\end{aligned}
\end{equation*}
This completes the proof of \eqref{GC}.

We now prove
\begin{equation} \label{regular1}
	\inf_{d(\theta, \theta_0)>\epsilon, \theta\in\mathbb{R}_C^d\times\mathcal{M}_C^\infty}	L_0(\theta)> L_0(\theta_0)
\end{equation}
with $\mathcal{M}_C^\infty=\cup_{i=1}^\infty\mathcal{M}_C^i$.
According to the equation (C.46) of \cite{BELLONI20194}, for any two scalars $a,b$, it holds that
\begin{equation}\label{abs}
	|a-b|-|a|=-b\left(\frac{1}{2}-\mathbf{1}\{a\leq0\}\right)+\int_{0}^{b}(\mathbf{1}\{a\leq t\}-\mathbf{1}\{a\leq 0\})\mathrm{d}t.
\end{equation}
For any $\theta\in\mathbb{R}_C^d\times \mathbf{conv}(\mathcal{M}_C^\infty)$, we denote
$
	\Lambda(\theta;V):=X^\top\beta+g(Z)$ and $\Lambda(\theta_0;V):=X^\top\beta_0+g_0(Z)$.
Taking $a=Y-\Lambda(\theta_0;V)$ and $b=\Lambda(\theta;V)-\Lambda(\theta_0;V)$ into \eqref{abs}, we have
\begin{equation}\label{regular3}
\begin{aligned}
		&L_0(\theta)-L_0(\theta_0)\\
	=&\mathbb{E}\left[-b\left(\frac{1}{2}-\mathbf{1}\{a\leq0\}\right)+\int_{0}^{b}(\mathbf{1}\{a\leq t\}-\mathbf{1}\{a\leq 0\})\mathrm{d}t\right]\\
		\overset{(\rm V)}{=}&\mathbb{E}\left[\int_{0}^{b}(\mathbf{1}\{a\leq t\}-\mathbf{1}\{a\leq 0\})\mathrm{d}t\right]\\
		=&\mathbb{E}_V\left[\mathbb{E}_Y\left[\int_{0}^{\Lambda(\theta;V)-\Lambda(\theta_0;V)}\mathbf{1}\{Y-\Lambda(\theta_0;V)\leq t\}-\mathbf{1}\{Y-\Lambda(\theta_0;V)\leq 0\}\mathrm{d}t\Bigg| V\right]\right]\\
		=&\mathbb{E}_V\left[\int_{0}^{\Lambda(\theta;V)-\Lambda(\theta_0;V)}F_{Y\mid V}(\Lambda(\theta_0;V)+t)-F_{Y\mid V}(\Lambda(\theta_0;V))\mathrm{d}t\right]\\
		\overset{}{=}&\mathbb{E}_V\left[\int_{0}^{\Lambda(\theta;V)-\Lambda(\theta_0;V)}tf_{Y\mid V}(\Lambda(\theta_0;V))+\frac{t^2}{2}f'_{Y\mid V}(\Lambda(\theta_0;V)+\bar{t}_{V,t})\mathrm{d}t\right]\\
		\overset{(\rm VI)}{\geq} & \frac{1}{2c_0}\mathbb{E}_V\left[ \left|\Lambda(\theta;V)-\Lambda(\theta_0;V)\right|^2\right]-\frac{d_0}{6}\mathbb{E}_V\left[ \left|\Lambda(\theta;V)-\Lambda(\theta_0;V)\right|^3\right]
\end{aligned}
\end{equation}
with $\bar{t}_{V,t}$ between $0$ and $t$. The equality ${(\rm V)}$ holds due to
\begin{equation*}
	\begin{aligned}
		&\mathbb{E}\left[-b\left(\frac{1}{2}-\mathbf{1}\{a\leq0\}\right)\right]\\
		=&\mathbb{E}_V\left[\mathbb{E}_Y\left[(\Lambda(\theta;V)-\Lambda(\theta_0;V))\left(\frac{1}{2}-\mathbf{1}\{Y-\Lambda(\theta_0;V)\leq 0\}\right)\Bigg| V\right]\right]\\
		=&\mathbb{E}_V\left[\mathbb{E}_Y\left[(\Lambda(\theta;V)-\Lambda(\theta_0;V))\left(\frac{1}{2}-\mathbf{1}\{Y-\Lambda(\theta_0;V)\leq 0\}\right)\Bigg| V\right]\right]\\
		=&\mathbb{E}_V\left[(\Lambda(\theta;V)-\Lambda(\theta_0;V))\mathbb{E}_Y\left[\left(\frac{1}{2}-\mathbf{1}\{\varepsilon\leq 0\}\right)\Bigg| V\right]\right]\\
		=&0,
	\end{aligned}
\end{equation*}
while the inequality ${(\rm VI)}$ holds due to Assumption (A\ref{assump: param space}) and (A\ref{assump: error pdf}).

Let $$\bar{q}(\theta):=\frac{\left(\frac{1}{c_0}\right)^\frac{3}{2}\mathbb{E}_V\left[ \left|\Lambda(\theta;V)-\Lambda(\theta_0;V)\right|^2\right]^\frac{3}{2}}{d_0\mathbb{E}_V\left[ \left|\Lambda(\theta;V)-\Lambda(\theta_0;V)\right|^3\right]}$$ and consider the case
$
	\left(\frac{1}{c_0}\mathbb{E}_V[ |\Lambda(\theta;V)-\Lambda(\theta_0;V)|^2]\right)^{\frac{1}{2}}\leq \bar{q}(\theta).
$
It holds that
\begin{equation*}
	d_0\mathbb{E}_V\left[ \left|\Lambda(\theta;V)-\Lambda(\theta_0;V)\right|^3\right]\leq \frac{1}{c_0}\mathbb{E}_V\left[ \left|\Lambda(\theta;V)-\Lambda(\theta_0;V)\right|^2\right].
\end{equation*}
Then we have
\begin{equation}\label{cor1}
	L_0(\theta)-L_0(\theta_0)\geq \frac{1}{3c_0}\mathbb{E}_V\left[ \left|\Lambda(\theta;V)-\Lambda(\theta_0;V)\right|^2\right]=\frac{1}{3c_0}d(\theta,\theta_0)^2.
\end{equation}

Next, we consider the case
$		\left(\frac{1}{c_0}\mathbb{E}_V\left[ \left|\Lambda(\theta;V)-\Lambda(\theta_0;V)\right|^2\right]\right)^{\frac{1}{2}}> \bar{q}(\theta)$.
Let $\tilde{\theta}=((1-\alpha)\beta+\alpha\beta_0, (1-\alpha)g+\alpha g_0)$
such that $	\left(\frac{1}{c_0}\mathbb{E}_V[ |\Lambda(\tilde{\theta};V)-\Lambda(\theta_0;V)|^2]\right)^{\frac{1}{2}}=\bar{q}(\theta)$. Then it holds that $	1-\alpha=\frac{\bar{q}(\theta)}{\sqrt{\frac{1}{c_0}\mathbb{E}_V[ |\Lambda(\theta;V)-\Lambda(\theta_0;V)|^2]}}.$
On the other hand, we have
\begin{equation}\label{regular4}
	L_0(\theta)-L_0(\theta_0)\geq \frac{L_0(\tilde{\theta})-L_0(\theta_0)}{1-\alpha}=\frac{\sqrt{\frac{1}{c_0}\mathbb{E}_V\left[ \left|\Lambda(\theta;V)-\Lambda(\theta_0;V)\right|^2\right]}}{\bar{q}(\theta)}(L_0(\tilde{\theta})-L_0(\theta_0)).
\end{equation}
Note that
\begin{equation*}
\begin{aligned}
		\bar{q}(\theta)=&\frac{\left(\frac{1}{c_0}\right)^\frac{3}{2}\mathbb{E}_V\left[ \left|\Lambda(\theta;V)-\Lambda(\theta_0;V)\right|^2\right]^\frac{3}{2}}{d_0\mathbb{E}_V\left[ \left|\Lambda(\theta;V)-\Lambda(\theta_0;V)\right|^3\right]}\\
		=&\frac{\left(\frac{1}{c_0}\right)^\frac{3}{2}\mathbb{E}_V\left[ \left|\Lambda(\tilde{\theta};V)-\Lambda(\theta_0;V)\right|^2\right]^\frac{3}{2}}{d_0\mathbb{E}_V\left[ \left|\Lambda(\tilde{\theta};V)-\Lambda(\theta_0;V)\right|^3\right]}\\
		=&\frac{\bar{q}^3(\theta)}{d_0\mathbb{E}_V\left[ \left|\Lambda(\tilde{\theta};V)-\Lambda(\theta_0;V)\right|^3\right]}.
\end{aligned}
\end{equation*}
Then we have $d_0\mathbb{E}_V\left[ \left|\Lambda(\tilde{\theta};V)-\Lambda(\theta_0;V)\right|^3\right]=\bar{q}^2(\theta)$. Furthermore, by \eqref{regular3}, it holds that
\begin{equation}\label{seperate}
\begin{aligned}
		L_0(\tilde{\theta})-L_0(\theta_0)\geq&\frac{1}{2c_0}\mathbb{E}_V\left[ \left|\Lambda(\theta;V)-\Lambda(\theta_0;V)\right|^2\right]-\frac{d_0}{6}\mathbb{E}_V\left[ \left|\Lambda(\theta;V)-\Lambda(\theta_0;V)\right|^3\right]
		=\frac{1}{3}\bar{q}^2(\theta).
\end{aligned}
\end{equation}
Substituting \eqref{seperate} into \eqref{regular4}, we have
\begin{equation}\label{cor2}
	L_0(\theta)-L_0(\theta_0)\geq \frac{\bar{q}(\theta)}{3\sqrt{c_0}}d(\theta,\theta_0).
\end{equation}
Combining \eqref{cor1} and \eqref{cor2}, we complete the proof of \eqref{regular1}.

We now finish the proof of the consistency. For the function $g_0$, let
\begin{equation*}
	g^*_N:=\underset{g\in\mathcal{M}_C^N}{\text{argmin}} \Vert g-g_0 \Vert_{L_2}\ \text{and}\  \theta^*_N:=(\beta_0,g^*_N),
\end{equation*}
while Schmidt-Hieber (2020) proves
\begin{equation}\label{dista}
	d(\theta^*_N,\theta_0)= O(r_N)\to 0, \ \text{as}~N\to\infty;
\end{equation}
see Equation (26) of \cite{10.1214/19-AOS1875} for details. Due to the definition of $d(\cdot,\cdot)$, it holds that
\begin{equation*}
	\begin{aligned}
		\left|L_0(\theta^*_N)-L_0(\theta_0)\right|=&\left|\mathbb{E}\left[\left|Y-\beta_0^\top X-g^*_N(Z)\right|\right]-\mathbb{E}\left[\left|Y-\beta_0^\top X-g_0(Z)\right|\right]\right|\\
		\leq& \mathbb{E}\left[\left|g_N^*(Z)-g_0(Z)\right|\right]\\
		\leq& d(\theta^*_N, \theta_0).
	\end{aligned}
\end{equation*}
Combining with \eqref{dista}, we have
\begin{equation}\label{o1}
	L_0(\theta_N^*)\leq L_0(\theta_0)+o(1).
\end{equation}
On the other hand, it is easy to show that
\begin{equation}\label{inf}
	\begin{aligned}
		L_N(\hat{\theta}_N)+\lambda_N\mathcal{J}_{N,M}(\hat{\theta}_N)\leq& 	L_N({\theta}_N^*)+\lambda_N\mathcal{J}_{N,M}(\theta^*_N)\\
		L_0(\hat{\theta}_N)+L_N(\hat{\theta}_N)+\lambda_N\mathcal{J}_{N,M}(\hat{\theta}_N)-L_0(\hat{\theta}_N)\leq &L_0(\theta^*_N)+ L_N({\theta}_N^*)+\lambda_N\mathcal{J}_{N,M}(\theta^*_N)-L_0(\theta^*_N).
	\end{aligned}
\end{equation}
Applying \eqref{GC} to \eqref{inf}, we have
\begin{equation*}
	\begin{aligned}
		L_0(\hat{\theta}_N)\leq& L_0(\theta^*_N) + |L_N(\hat{\theta}_N)+\lambda_N\mathcal{J}_{N,M}(\hat{\theta}_N)-L_0(\hat{\theta}_N)|+\left|L_N({\theta}_N^*)+\lambda_N\mathcal{J}_{N,M}(\theta^*_N)-L_0(\theta^*_N)\right|\\
		\leq & L_0(\theta^*_N)+2	\mathop{\sup}_{\theta \in \mathbb{R}^{d}_{C}\times\mathcal{M}_C^N}|L_{N}(\theta)+\lambda_N\mathcal{J}_{N,M}(\theta)-L_{0}(\theta)|\\
		\overset{\eqref{GC}}{\leq} & L_0(\theta^*_N)+o_p(1).
	\end{aligned}
\end{equation*}

Combining with \eqref{o1}, it holds that
\begin{equation}\label{regular2}
			L_0(\hat{\theta}_N)\leq L_0(\theta_0)+o(1)+o_p(1)\leq L_0(\theta_0)+o_p(1).
\end{equation}
If there exists $\epsilon_1>0$ such that$	P(d(\hat{\theta}_N,\theta_0)>\epsilon_1)>0,$
then, based on \eqref{regular1},
\begin{equation*}
	P(L_0(\hat{\theta}_N)>L_0(\theta_0)+\epsilon_2)>	P(d(\hat{\theta}_N,\theta_0)>\epsilon_1)>0
\end{equation*}
holds for some $\epsilon_2>0$, which
contradicts with \eqref{regular2}. This completes the proof of \eqref{consi}.
%

We now prove that
\begin{equation}\label{conv_rate}
	d(\hat{\theta}_N,\theta_0)=O_p(r_N\log^2 N+\lambda_N),
\end{equation}
following the line of Theorem 3.4.6 of \cite{Vaart2023}. We set the parameters $({\{\theta_n\}_{n=1}^\infty},\{\theta_{n,0}\}_{n=1}^\infty,c,\{\underline{\delta}_n\}_{n=1}^{\infty},\{\lambda_n\}_{n=1}^\infty)$ of Theorem 3.4.6 of \cite{Vaart2023} to our counterparts $(\{\theta_N^*\}_{N=1}^\infty,\{\theta_0 \}_{N=1}^\infty,0,\{0\}_{N=1}^\infty,\allowbreak \{\lambda_N\}_{N=1}^\infty)$, respectively. Furthermore, we
write $R:= 2H^2(L+1)$ and
\begin{equation}\label{A_delta}
	\mathcal{A}_{\delta}^N = \{\theta \in \mathbb{R}^{d}_{C}\times \mathcal{M}_C^N~|~  d(\theta, \theta_0)\le \delta \}.
\end{equation}
By Theorem 3.4.6 of \cite{Vaart2023}, it suffices to verify that, for any $\delta>0$,
\begin{align}
	&\delta^2\lesssim\inf_{\theta\in \mathbb{R}_C^d\times \mathcal{M}_C^N\colon \frac{\delta}{2}< d(\theta,\theta_0)\leq \delta} L_0(\theta)-L_0(\theta_0),\label{donsker1}\\
	&	\mathbb{E}\left[ \mathop{\sup}_{\substack{\theta \in\mathcal{A}_{\delta}^N,\\ \mathcal{J}_{N,M}(\theta)<\delta/\lambda_N} }\sqrt{N}\left|(L_N-L_0)(\theta_N^*)-(L_N-L_0)(\theta)\right|\right]\lesssim\phi_N(\delta)\label{donsker2}
\end{align}
with  ${\phi_N(\delta)=\delta\sqrt{s\log\frac{R}{\delta}}+ \frac{s}{\sqrt{N}}\log \frac{R}{\delta}}$.

Indeed, for every ${\tilde{M}}>0$ there exists a constant $\gamma_{\tilde{M}}>0$ such that $	\mathbb{E}_{\varepsilon}\left[|\varepsilon|\right]-\mathbb{E}_{\varepsilon}\left[|\varepsilon+\mu|\right]\leq -\gamma_{\tilde{M}} \left|\mu\right|^2$
for $|\mu|\leq \tilde{M}$.
Then for any $(\beta,g) \in\mathbb{R}_C^d\times \mathcal{M}_C^\infty$, we have
\begin{equation}\label{taylor2}
	L_0(\theta_0)-L_0(\theta)\lesssim -\gamma_{\tilde{M}}d^2(\theta,\theta_0)
\end{equation}
with ${\tilde{M}}:=\sup_{(\beta,g) \in \mathbb{R}^{d}_{C}\times\mathcal{M}_C^\infty} 2\Vert\beta\Vert_\infty+2\Vert g\Vert_\infty$.
Then, \eqref{donsker1} holds by taking $\frac{\delta}{2}< d(\theta,\theta_0)$ into \eqref{taylor2}.

We now verify \eqref{donsker2}.
Denote $\rho(\theta;U):=|Y-\beta^\top X-g(Z)|,\forall U= (X,Y,Z)\in\mathbb{R}\times [0,1]^{d+l}$ and $\mathcal{B}_\delta^N=\{\rho(\theta_N^*;U)-\rho(\theta;U)~|~\theta\in\mathcal{A}_{\delta}^N\}.$ 
For any $\theta,\theta_1\in\mathcal{A}_{\delta}^N$, we have 
$\mathbb{E}|\rho(\theta;U)-\rho(\theta_1;U)|^2\le 4d^2(\theta,\theta_1)$.  
Lemma 5 of \cite{10.1214/19-AOS1875} then implies that
\begin{equation}\label{inq_num2}
	\begin{aligned}
			\log\left(N_{[~]}(\epsilon,\mathcal{B}_{\delta}^N,L^{2}(P))\right)\leq &\log\left( N_{[~]}(\epsilon,\mathcal{B}_\delta^N,\Vert\cdot\Vert_\infty)\right)\\
			\overset{(\rm VII)}{\leq} & \log \left(N(\epsilon,\mathcal{B}_\delta^N,\Vert\cdot\Vert_\infty)\right)\\
			\leq & \log\left( N(\epsilon,\mathcal{F}_N,\Vert\cdot\Vert_\infty)\right)\\
			{\leq} & \log\left(K_\epsilon N\left(\frac{\epsilon}{2}, \mathcal{M}_C^N,\Vert\cdot\Vert_\infty\right)\right)\quad (\text{by}\  \eqref{covering num})\\
			{\lesssim}& \log(K_\epsilon)+s\log\frac{R}{\epsilon}  (\text{By} \eqref{lemma 5})
	\end{aligned}
\end{equation}
where ${N}_{[~]}(\epsilon,\mathcal{B}_{\delta}^N,L^{2}(P))$ ($N_{[~]}(\epsilon,\mathcal{B}_\delta^N,\Vert\cdot\Vert_\infty)$) is the bracket number of $\mathcal{B}_{\delta}^N$ with $L^{2}(P)$ norm ($L^\infty$ norm). The inequality $(\rm VII)$ holds due to Page 132 of \cite{Vaart2023}. Henceforth, it follows that
\begin{equation*}
	J_{[~]}(\delta,\mathcal{B}_{\delta}^N)=\int_{0}^{\delta}\sqrt{1+ \log\left({N}_{[~]}(\epsilon,\mathcal{B}_{\delta}^N,L^{2}(P))\right)}\mathrm{d}\epsilon\lesssim {\delta\sqrt{s\log\frac{R}{\delta}}},
\end{equation*}
where the last inequality holds by noticing
\begin{equation}
	\begin{aligned}\label{inte_ineq}
		\int_0^\delta \sqrt{\log \frac{R}{\epsilon}} \mathrm{d} \epsilon=\delta \sqrt{\log \frac{R}{\delta}}+\frac{R \sqrt{\pi}}{2} \operatorname{erfc}\left(\sqrt{\log \frac{R}{\delta}}\right)\lesssim \delta\sqrt{\log\frac{R}{\delta}}
	\end{aligned}
\end{equation}
with $\operatorname{erfc}(x)=\frac{2}{\sqrt{\pi}} \int_x^{\infty} e^{-t^2} \mathrm{d} t$.
By Lemma 3.4.2 of \cite{Vaart2023}, we conclude that 
\begin{equation*}
	\begin{aligned}
		&\mathbb{E} \left[\mathop{\sup}_{\beta\in\mathcal{A}_{\delta}^N, \mathcal{J}_{N,M}(\theta)<\delta/\lambda_N}\sqrt{N}\left|(L_N-L_0)(\theta_N^*)-(L_N-L_0)(\theta)\right|\right]\\
		=& \mathbb{E}\left[ \mathop{\sup}_{\beta\in\mathcal{A}_{\delta}^N, \mathcal{J}_{N,M}(\theta)<\delta/\lambda_N}\left|\mathbb{G}_N(\rho(\theta_N^*; U)-\rho(\theta;U))\right|\right]\\
		\lesssim& J_{[~]}(\delta,\mathcal{B}_{\delta}^N)\Big\{ \frac{J_{[~]}(\delta,\mathcal{B}_{\delta}^N)}{\delta^2\sqrt{N}}+1\Big\}\\
		=&\phi_N(\delta).
	\end{aligned}
\end{equation*}

Setting $\delta_N = \eta_N=r_N\log^2N$ in Theorem 3.4.6 of \cite{Vaart2023}, it can be verified that 
\begin{equation}\label{convex_sqrt}
	\frac{1}{\eta_N^2}\phi_N(\eta_N)\lesssim\sqrt{N}~\text{and}~L_N(\hat{\theta}_N)+\lambda_N\mathcal{J}_{N,M}(\hat{\theta}_N)\le L_{N}(\theta^*_N)+\lambda_N\mathcal{J}_{N,M}(\theta^*_N).
\end{equation}
Then, by Theorem 3.4.6 of \cite{Vaart2023}, we obtain $d(\hat{\theta}_N,\theta_0)=O_p(r_N\log^2 N+\lambda_N)$ and $\mathcal{J}_{N,M}(\hat{\theta}_N)= O_p(\eta_N/\lambda_N+1)$. 

Furthermore, by Assumption (A\ref{assump: error pdf}),
\begin{equation}
\begin{aligned}\label{derive_conve}
    d^2(\hat{\theta}_N,\theta_0)&=\mathbb{E}\{X^\top(\hat{\beta}_N-\beta_0) + \hat{g}_N(Z)-g_0(Z)\}^2\\
    &=\mathbb{E}[\{(X-\mathbb{E}[X|Z])^\top (\hat{\beta}_N-\beta_0) + (\hat{\beta}_N-\beta_0)^\top \mathbb{E}[X|Z]+ \hat{g}_N(Z)-g_0(Z)\}^2]\\
    &=\mathbb{E}[\{ (X-\mathbb{E}[X|Z])^\top(\hat{\beta}_N-\beta_0)\}^2]\\
    &~~~+\mathbb{E}[\{(\hat{\beta}_N-\beta_0)^\top \mathbb{E}[X|Z] + \hat{g}_N(Z)-g_0(Z)\}^2].
\end{aligned}
\end{equation}
Since the matrix $\mathbb{E}[\{X-\mathbb{E}[X|Z]\}\{X-\mathbb{E}[X|Z]\}^\top]$ is positive definite,
 it follows that $\|\hat{\beta}_N-\beta_0\|_\infty=O_p(r_N\log^2 N+\lambda_N)$ and thus $\|\hat{g}_N-g_0\|_{L^2(P)}=O_p(r_N\log^2 N+\lambda_N)$. This completes the proof.\\
\end{proof}


 \begin{theorem}\label{equi_cont}
	Let $\mathcal{F}_{\delta_N}:=\{f-g:f,g\in\mathcal{F}_N,\  \Vert f-g\Vert_{L^2(P)}\leq \delta_N\}$ with $\delta_N=O(r_N\log^2 N)$. Under Assumptions (A\ref{assump: param space})-(A\ref{assump:rate_lambda}), it holds that
	\begin{align*}
	\Vert\mathbb{G}_N\Vert_{\mathcal{F}_{\delta_N}}\xrightarrow{p}0,\quad
	\mathbb{G}_N(\hat{f}_N) \leadsto \mathcal{N}(0, \Sigma)
	\end{align*}
 with $\hat{f}_N = |Y - \hat{\beta}_N^\top X - \hat{g}_N(Z)|$ and $\Sigma = \mathrm{Var}_U(|Y - \beta_0^\top X - g_0(Z)|)$.
\end{theorem}
\begin{proof}
	By the Markov inequality and Lemma 2.3.1 of \cite{Vaart2023}, it holds that
	   \begin{align}\label{empir_pro}
	   	P(\Vert\mathbb{G}_N\Vert_{\mathcal{F}_{\delta_N}}>x)\leq& \frac{2}{x} \mathbb{E}_{U,e}\left\Vert\frac{1}{\sqrt{N}}\sum_{i=1}^{N}e_i f(U_i)\right\Vert_{\mathcal{F}_{\delta_N}}\nonumber\\
	   	=&\frac{2}{x}\mathbb{E}\mathbb{E}_e\left\Vert\frac{1}{\sqrt{N}}\sum_{i=1}^{N}e_i f(U_i)\right\Vert_{\mathcal{F}_{\delta_N}}\nonumber\\
	   {\lesssim}&\frac{2}{x} \mathbb{E}\left[\int_{0}^{\infty}\sqrt{\log D(\epsilon,\mathcal{F}_{\delta_N},L^2(\mathbb{P}_N))} \mathrm{d}\epsilon\right]\nonumber\quad {(\text{by Corollary 2.2.9 of \cite{Vaart2023}})}\\
	   	\overset{}{\lesssim}&\frac{2}{x} \mathbb{E}\left[\int_{0}^{\infty}\sqrt{\log N(\epsilon,\mathcal{F}_{\delta_N},L^2(\mathbb{P}_N))} \mathrm{d}\epsilon\right]\\
	   	{=}& \frac{2}{x} \mathbb{E}\left[\int_{0}^{\Xi_N^2}\sqrt{\log N(\epsilon,\mathcal{F}_{\delta_N},L^2(\mathbb{P}_N))} \mathrm{d}\epsilon\right]\nonumber \quad (\text{with} \ \Xi_N^2:=\left\Vert\frac{1}{N}\sum_{i=1}^{N}f^2(U_i)\right\Vert_{\mathcal{F}_{\delta_N}})\\
	   	\overset{(\rm I)}{\lesssim}&\frac{2}{x} \mathbb{E}\left[\int_{0}^{\Xi_N^2}\sqrt{\log N(\epsilon,\mathcal{F}_{N},\Vert\cdot\Vert_\infty)} \mathrm{d}\epsilon\right]\nonumber\\
	   	\overset{}{\lesssim}& \frac{2}{x} \mathbb{E}\left[\int_{0}^{\Xi_N^2}\sqrt{ s\log\left(\frac{R}{\epsilon}\right)} \mathrm{d}\epsilon\right],\nonumber
	   \end{align}
 where according to $N(\epsilon,\mathcal{F}_{\delta_N},L^2(\mathbb{P}_N))\lesssim N^2(\epsilon,\mathcal{F}_N,L^2(\mathbb{P}_N))\leq N^2(\epsilon,\mathcal{F}_N,\Vert \cdot\Vert_\infty)$, the inequality ($\rm I$) holds.

Note that
\begin{equation}\label{triangle}
	\begin{aligned}
		\Xi_N^2=\!\Vert\mathbb{P}_N f^2\Vert_{\mathcal{F}_{\delta_N}}=\Vert Pf^2\Vert_{\mathcal{F}_{\delta_N}}+\Vert\mathbb{P}_N f^2-P f^2\Vert_{\mathcal{F}_{\delta_N}}\leq\delta_N^2+\Vert\mathbb{P}_N f^2-P f^2\Vert_{\mathcal{F}_{\delta_N}}.
	\end{aligned}
\end{equation}
Taking \eqref{triangle} into \eqref{empir_pro}, we obtain
\begin{equation}
	\begin{aligned}
			P(\Vert\mathbb{G}_N\Vert_{\mathcal{F}_{\delta_N}}>x)=&\frac{2}{x} \mathbb{E}\left[\int_{0}^{\delta_N^2+\Vert\mathbb{P}_N f^2-P f^2\Vert_{\mathcal{F}_{\delta_N}}}\sqrt{ s\log\left(\frac{R}{\epsilon}\right)} \mathrm{d}\epsilon\right]\\
			\lesssim& \frac{2}{x}\mathbb{E} \left[(\delta_N^2+\Vert\mathbb{P}_N f^2-P f^2\Vert_{\mathcal{F}_{\delta_N}})\sqrt{s\log\left(\frac{R}{\delta_N^2+\Vert\mathbb{P}_N f^2-P f^2\Vert_{\mathcal{F}_{\delta_N}}}\right)}\right]\\
			\leq & \frac{2}{x}\mathbb{E}\left[\delta_N^2\sqrt{s \log\left(\frac{R}{\delta_N^2}\right)}\right]+\frac{2}{x}\mathbb{E}\left[ \Vert\mathbb{P}_N f^2-P f^2\Vert_{\mathcal{F}_{\delta_N}}\sqrt{s \log\left(\frac{R}{\delta_N^2}\right)}\right]\\
			{\leq} &\frac{2}{x}\widetilde{O}(r_N^3\sqrt{N})+\frac{2}{x} \widetilde{O}(r_N^2\sqrt{N}) \quad (\text{by Assumption}\ (\text{A}\ref{assump:rate_lambda}))\\
			{=}&o(1) \hspace{3.9cm}  (\text{by Assumption (A\ref{assump: covariates})})
	\end{aligned}
\end{equation}
Based on Slutsky's theorem, we have $\mathbb{G}_N(\hat{f}_N) \leadsto \mathcal{N}(0, \Sigma)$ with $\hat{f}_N = |Y - \hat{\beta}_N^\top X - \hat{g}_N(Z)|$ and $\Sigma = \text{Var}_U(|Y - \beta_0^\top X - g_0(Z)|)$.
\end{proof}

\begin{theorem}\label{thm_clt}
Under the assumptions (A\ref{assump: param space})-(A\ref{assump: regular}), it holds that
\begin{equation*}
  \sqrt{N}(\hat{\beta}_N-\beta_0)\leadsto \mathcal{N}(0,\Sigma_2^{-1}\Sigma_1\Sigma_2^{-1})
\end{equation*}
\end{theorem}

\begin{proof}

For $\hat{\theta}_N=(\hat{\beta}_N, \hat{g}_N)$, we introduce the following notations:
${\xi}={\beta}-\beta_0,$ $\hat{\xi}_N=\hat{\beta}_N-\beta_0,$ ${h}(Z)={g}(Z)-g_0(Z)+(\beta-\beta_0)^\top \boldsymbol{\varphi}^*(Z),$ $\hat{h}_N(Z)=\hat{g}_N(Z)-g_0(Z)+(\hat{\beta}_N-\beta_0)^\top \boldsymbol{\varphi}^*(Z)$ and $\tilde{X}=X-\boldsymbol{\varphi}^*(Z)$.  These imply that 
\begin{equation*}
	\frac{1}{N}\sum_{i=1}^{N} \vert Y_i-\beta^\top X_i-g(Z_i)\vert=
	\frac{1}{N}\sum_{i=1}^{N} \vert\varepsilon_i-\xi^\top \tilde{X}_i-h(Z_i)\vert.
\end{equation*}
Denote $M_N(\xi,h)=\frac{1}{N}\sum_{i=1}^{N} \vert\varepsilon_i-\xi^\top \tilde{X}_i-h(Z_i)\vert$, and we may calculate the subgradient of the loss function $M_N$ at $\xi$ as 
\begin{equation*}
	\partial_{\xi} M_N(\xi, h) = \frac{1}{N}\sum_{i=1}^{N}  \left(-\mathbf{sign^*}(\varepsilon_i-\xi^\top \tilde{X}_i-h(Z_i))\tilde{X}_i\right)
\end{equation*}
with
 Clearly, letting $\Psi_N(\xi,h):=\mathbb{P}_N\psi(\xi,h)$ with
$\psi(\xi,h)=-\mathbf{sign}(\varepsilon-\xi^\top \tilde{X}-h(Z))\tilde{X}$, we have $\Psi_N(\xi,h)\in \partial_{\xi} M_N(\xi, h)$. We further denote
\begin{equation*}
	\begin{aligned}
		&(\xi_0,h_0(Z))=(0,0)\in\mathbb{R}^{d}\times L^2(P),\\
		&\Psi_0(\xi,h)=\mathbb{E}\psi(\xi,h),\\
		&\widetilde{\mathcal{A}}_{\delta}^N=\{(\xi,h)~|~\xi={\beta}-\beta_0,{h}(Z)={g}(Z)-g_0(Z)+(\beta-\beta_0)^\top \boldsymbol{{\varphi}}^*(Z),(\beta,g)\in\mathcal{A}_{\delta}^N\},\\
		&\mathcal{C}_{\delta}^N=\{\psi(\xi,h)-\psi(\xi_0,h_0)~|~(\xi,h)\in \widetilde{\mathcal{A}}_{\delta}^N\},
	\end{aligned}
\end{equation*}
for the convenience of the following discussions.  Although $(\xi_0,h_0)=(0,0)$ is constant, we still introduce such notation to articulate the fields ($(\beta, g)$ or $(\xi,h)$) to analysis $\Psi_0$ and $\Psi_N$.

By analogy to the proof of Theorem \ref{thm1}, we have
\begin{equation*}
	\log N_{[~]}(\epsilon, {\mathcal{A}}_\delta^N, \Vert\cdot\Vert_\infty)\lesssim s\log \frac{R}{\epsilon}.
\end{equation*}
Let $\{[l_i, u_i]\colon l_i=(\beta_{i;l}, g_{i;l}),u_i=(\beta_{i;u}, g_{i;u}), i=1,\dots, K \}$ be the $\epsilon$-brackets of ${\mathcal{A}}_\delta^N$ with $K=N(\epsilon, {\mathcal{A}}_\delta^N, L^\infty(P))$, and for any $(\beta,g)\in{\mathcal{A}}_\delta^N$, without loss of generality, we assume that $[l_1, u_1]$ is the $\epsilon$-bracket of $(\beta, g)$. We may notice that
\begin{equation*}
	\begin{aligned}
		&\int \left\vert \mathbf{sign}\left(\varepsilon-\xi_{1;l}^\top \tilde{X}-h_{1;l}(Z)\right) - \mathbf{sign}\left(\varepsilon-\xi_{1;u}^\top \tilde{X}-h_{1;u}(Z)\right)\right\vert^2 \mathrm{d}P_{\varepsilon,\widetilde{X},Z}\\
		\overset{\rm (\Delta_1)}{=}& \int \left\vert \mathbf{sign}\left(Y-\beta_{1;l}^\top {X}-g_{1;l}(Z)\right) - \mathbf{sign}\left(Y-\beta_{1;u}^\top {X}-g_{1;u}(Z)\right)\right\vert^2 \mathrm{d}P_{Y,{X},Z}\\
		\overset{\rm (\Delta_2)}{\leq} & 4 P\left(Y-\beta_{1;l}^\top {X}-g_{1;l}(Z)\geq 0, Y-\beta_{1;u}^\top {X}-g_{1;u}(Z)< 0\right)\\
		\overset{\rm (\Delta_3)}{\lesssim} & \sup\limits_{\tau\in\mathbb{R}} P(Y\in [\tau,\tau+(C+1)\epsilon])\overset{\rm (\Delta_4)}{\lesssim} \epsilon,
	\end{aligned}
\end{equation*}
where we denote $\xi_{1;\nu}=\beta_{1;\nu}-\beta_0$ and ${h}_{1;\nu}(Z)={g}_{1;\nu}(Z)-g_0(Z)+(\beta_{1;\nu}-\beta_0)^\top \boldsymbol{{\varphi}}^*(Z)$ in equality ($\mathrm{\Delta_1}$) with $\nu\in\{l,u\}$. Moreover, inequality ($\mathrm{\Delta_2}$) holds due to $\mathbf{sign}(Y-\beta_{1;l}^\top {X}-g_{1;l}(Z))\geq \mathbf{sign}(Y-\beta_{1;u}^\top {X}-g_{1;u}(Z))$, and inequalities ($\mathrm{\Delta_3}$) and  ($\mathrm{\Delta_4}$)  hold by  Assumption  (A\ref{assump: covariates}) and (A\ref{assump:rate_lambda}). Hence, we can deduce that
\begin{equation*}
	\log N_{[~]}(\epsilon,\mathcal{C}_{\delta}^N,L^{2}(P))\lesssim s\log \frac{R}{\epsilon};
\end{equation*}
thus for any $\delta>0,$
\begin{equation*}
	J_{[~]}(\delta,\mathcal{C}_{\delta}^N)=\int_{0}^{\delta}\sqrt{1+ N_{[~]}(\epsilon,\mathcal{C}_{\delta}^N,L^{2}(P))}d\epsilon\lesssim \delta\sqrt{s\log\frac{R}{\delta}}.
\end{equation*}

	Let $\delta_N=O(r_N\log^2N+\lambda_N)$, it follows
	\begin{equation*}
		\begin{aligned}
			&\mathbb{E}\left\{ \mathop{\sup}_{(\xi,h)\in\mathcal{C}_{\delta_N}^N}\big\|\sqrt{N}[(\Psi_N-\Psi_0)(\xi,h)-(\Psi_N-\Psi_0)(\xi_0,h_0)]\big\|\right\} \\=& \mathbb{E}\left\{ \mathop{\sup}_{(\xi,h)\in\mathcal{C}_{\delta_N}^N}\big\|\sqrt{N}(\mathbb{P}_N-P)[\psi_{\tau}(\xi,h)-\psi_{\tau}(\xi_0,h_0)]\big\|\right\}\\
			\overset{(\Delta_5)}{\lesssim}& J_{[~]}({\delta_N},\mathcal{C}_{\delta_N}^N)\left\{ \frac{J_{[~]}({\delta_N},\mathcal{C}_{\delta_N}^N)}{\delta_N^2\sqrt{N}}+1\right\}\\
			=&o(1),
		\end{aligned}
	\end{equation*}
where the inequality ($\Delta_5$) holds by Theorem 2.14.18$^\prime$ of \cite{Vaart2023}.
Since $\|\hat{\xi}_N\|\vee\|\hat{h}_N\|_{L^2([0,1]^d)}=O_{P}(r_N\log^2 N+\lambda_N),$ we have
\begin{equation*}
	\mathbb{E}\left\|\sqrt{N}[(\Psi_N-\Psi_0)(\hat{\xi}_N,\hat{h}_N)-(\Psi_N-\Psi_0)(\xi_0,h_0)]\right\|=o(1),
\end{equation*}
or, written alternatively,
\begin{equation}\label{proof: raw score}
	\sqrt{N}\{\Psi_0(\hat{\xi}_N,\hat{h}_N)+\Psi_N(\xi_0,h_0)\}=\sqrt{N}\{\Psi_N(\hat{\xi}_N,\hat{h}_N)+\Psi_0(\xi_0,h_0)\} + o_p(1).
\end{equation}

  Let $\tilde{Y}_{i,N}=\varepsilon_i-\hat{h}_{N}(Z_i),~i=1,\ldots,N.$ Then $\hat{\xi}_N$ is the minimizer of $M_N^*(\xi)=\frac{1}{N}\sum_{i=1}^{N}|\tilde{Y}_{i,N}-\xi^\top \tilde{X}_i|$ with respect to $\xi$ and
  \begin{equation}\label{subgradien}
  	\Psi_N(\hat{\xi}_N,\hat{h}_N)=-\frac{1}{N}\sum_{i=1}^{N}\textbf{sign}(\tilde{Y}_{i,N}- \hat{\xi}^\top_N \tilde{X}_{i})\tilde{X}_{i}.
  \end{equation}
Since $M_N^{*}$ is a continuous piecewise function of $\xi,$ it follows that the limiting subgradient is bounded by the difference between the right and left derivatives. Thus, we have
\begin{equation*}
\begin{aligned}
		0\overset{(\Delta_6)}{\in}&{\partial}_{\xi}\left(M_N^*+\lambda_N\mathcal{J}_{N,M}\right)\big|_{(\xi,h)=(\hat{\xi}_N,\hat{h}_N)}\\
	\overset{(\Delta_7)}{\subseteq}&{\partial}_{\xi} M_N^*\big|_{(\xi,h)=(\hat{\xi}_N,\hat{h}_N)}+\lambda_N{\partial}_{\xi} \mathcal{J}_{N,M}\big|_{(\xi,h)=(\hat{\xi}_N,\hat{h}_N)}.
\end{aligned}
\end{equation*}
where $(\Delta_6)$ holds by Theorem 10.1 of \cite{Rockafellar1998}, and $(\Delta_7)$ holds by Exercise 10.10 and Equation 10(6) of \cite{Rockafellar1998}.
Thus, we have
	\begin{align*}
		&\partial_{(\xi,h)} \mathcal{J}_{N,M}\\
		&{=}\frac{\partial\theta}{\partial (\xi,h)}\Big|_{(\xi,h)=(\hat{\xi}_N,\hat{h}_N)}^*\partial_{\theta} \mathcal{J}_{N,M}\Big|_{(\beta,g)=(\hat{\beta}_N,\hat{g}_N)} \quad {(\text{by Proposition 1.37 of \cite{mordukhovich2024second}})}\\
		&\overset{(\Delta_8)}{=}\frac{\partial\theta}{\partial (\xi,h)}\Big|_{(\xi,h)=(\hat{\xi}_N,\hat{h}_N)}^*\left[\begin{array}{l}
			\partial_{\beta}\mathcal{J}_{N,1}\Big|_{(\beta,g)=(\hat{\beta}_N,\hat{g}_N)}\\ \partial_{g}\mathcal{J}_{N,2}\Big|_{(\beta,g)=(\hat{\beta}_N,\hat{g}_N)}
		\end{array}\right]\\
		&{=}\left[\begin{array}{cc}
			I& 0\\
			\boldsymbol{\varphi}^*(Z)& \mathbf{I}
		\end{array}\right]^*\left[\begin{array}{l}
	\partial_{\beta}\mathcal{J}_{N,1}\Big|_{(\beta,g)=(\hat{\beta}_N,\hat{g}_N)}\\ \partial_{g}\mathcal{J}_{N,2}\Big|_{(\beta,g)=(\hat{\beta}_N,\hat{g}_N)}
\end{array}\right] \quad {(\text{by the definition of $\xi, h$})}\\
&=\left[\begin{array}{l}
		\partial_{\beta} \mathcal{J}_{N,1}\Big|_{(\beta,g)=(\hat{\beta}_N,\hat{g}_N)}+\underbrace{\left(\partial_{g}\mathcal{J}_{N,2}\Big|_{(\beta,g)=(\hat{\beta}_N,\hat{g}_N)}(\boldsymbol{\varphi}^*_1(Z)),\dots,\partial_{g}\mathcal{J}_{N,2}\Big|_{(\beta,g)=(\hat{\beta}_N,\hat{g}_N)}(\boldsymbol{\varphi}_d^*(Z))\right)^\top}_{\mathbf{G}_N}\\
		\partial_{g}\mathcal{J}_{N,2}\Big|_{(\beta,g)=(\hat{\beta}_N,\hat{g}_N)}
	\end{array}\right]
	\end{align*}
where $(\Delta_8)$ holds by Proposition 10.5 of \cite{Rockafellar1998} since $\mathcal{J}_{N,M}(\theta)=\mathcal{J}_{N,1}(\beta)+\mathcal{J}_{N,2}(g)$ and $(\mathbf{d}\mathcal{J}_{N,1}(\hat{\beta}_N)(0),\mathbf{d}\mathcal{J}_{N,2}(\hat{g}_N)(0))=0$. Here, $\mathcal{M}_C^N$ can be embedded in $L^2(\mathbf{m})$ space with Lebesgue measure $\mathbf{m}$, and $\partial_g\mathcal{J}_{N,2}\Big|_{(\beta,g)=(\hat{\beta}_N,\hat{g}_N)}\in (L^2(\mathbf{m}))^*$ is the differential operator with norm $\|\cdot\|_{L^2(\mathbf{m})}$. By Riesz representation theorem, we can represent $\partial_g\mathcal{J}_{N,2}\Big|_{(\beta,g)=(\hat{\beta}_N,\hat{g}_N)}(\boldsymbol{\varphi}^*_i(Z))=\int f_{\mathcal{J}}(Z)\boldsymbol{\varphi}^*_i(Z)\mathbf{m}(\mathrm{d}Z), i=1,\dots, d$ with a potential element $f_{\mathcal{J}}(Z)\in L^2(\mathbf{m})$.
Finally, by Assumption (A\ref{assump: regular}) and Corollary 3.44 of \cite{mordukhovich2006variational}, we have
\begin{equation*}
	\begin{aligned}
		\partial_{\xi} \mathcal{J}_{N,M}\Big|_{(\xi,h)=(\hat{\xi}_N,\hat{h}_N)}\subseteq\partial_{\beta} \mathcal{J}_{N,1}\Big|_{(\beta,g)=(\hat{\beta}_N,\hat{g}_N)}+\mathbf{G}_N.
	\end{aligned}
\end{equation*}
Let $\mathcal{I}_0:=\{i|\tilde{Y}_{i,N}=\hat{\xi}_N^\top\tilde{X}_i\}$, it holds that
\begin{equation}\label{optimality}
	0=\frac{1}{N}\left[\sum_{i=1,i\in\mathcal{I}_0}^{N} g_i\tilde{X}_i-\sum_{i=1,i\notin\mathcal{I}_0}^{N}\textbf{sign}(\tilde{Y}_{i,N}- \hat{\xi}^\top_N \tilde{X}_{i})\tilde{X}_{i}\right]+\lambda_N(\mathcal{Q}_1+\mathcal{Q}_2),
\end{equation}
where $g_i\in[-1,1], i\in\mathcal{I}_0$ are some subgradients of the absolute value function and $\mathcal{Q}_1\in\partial_\beta \mathcal{J}_{N,1}\Big|_{(\beta,g)=(\hat{\beta}_N,\hat{g}_N)}$ and $\mathcal{Q}_2\in\mathbf{G}_N$. According to Assumption (A\ref{assump: regular}), we have $\lambda_N\|\mathcal{Q}_1+\mathcal{Q}_2\|=o_p(\frac{1}{\sqrt{N}})$.
Taking \eqref{optimality} into \eqref{subgradien}, it holds that
  	\begin{align*}
  		\left\|\frac{\partial M_N^*(\xi)}{\partial \xi}\big|_{\xi=\hat{\xi}_N}\right\|_1\leq& \frac{1}{N}\left[\sum_{i=1,i\in\mathcal{I}_0}^{N}\left\|\textbf{sign}(\tilde{Y}_{i,N}- \hat{\xi}^\top_N \tilde{X}_{i})\tilde{X}_{i}\right\|+\sum_{i=1,i\in\mathcal{I}_0}^{N} \left\|g_i\tilde{X}_i\right\|\right]+\lambda_N\left\|\mathcal{Q}_1+\mathcal{Q}_2\right\|\\
  		\leq&\frac{1}{N}\left[\sum_{i=1,i\in\mathcal{I}_0}^{N}\|\tilde{X}_i\|+\|g_i\|\|\tilde{X}_i\|\right]+\lambda_N\left\|\mathcal{Q}_1+\mathcal{Q}_2\right\|\\
  		\leq&\frac{2}{N}\sum_{i=1,i\in\mathcal{I}_0}^{N}\|\tilde{X}_{i}\|+\lambda_N\left\|\mathcal{Q}_1+\mathcal{Q}_2\right\|\\
  		\leq&\frac{2}{N}\sum_{i=1}^{N}\mathbf{1}\{\tilde{Y}_{i,N}=\hat{\xi}^\top_N \tilde{X}_{i}\}\|\tilde{X}_{i}\|+\lambda_N\left\|\mathcal{Q}_1+\mathcal{Q}_2\right\|\\
  		\leq&\Big\{2\sum_{i=1}^{N}\mathbf{1}\{\tilde{Y}_{i,N}=\hat{\xi}^\top_N \tilde{X}_{i}\}\Big\}\max_{i=1,\ldots,N}\Big(\frac{\|\tilde{X}_{i}\|}{N}\Big)+\lambda_N\left\|\mathcal{Q}_1+\mathcal{Q}_2\right\|\\
  		\overset{(\Delta_9)}{\leq}& d \max_{i=1,\ldots,N}\Big(\frac{\|\tilde{X}_{i}\|}{N}\Big)+\lambda_N\left\|\mathcal{Q}_1+\mathcal{Q}_2\right\|\\
  		\overset{}{\leq}& d \max_{i=1,\ldots,N}\Big(\frac{\|\tilde{X}_{i}\|}{N}\Big)+o_p\left(\frac{1}{\sqrt{N}}\right)\\
  		=&o_p\left(\frac{1}{\sqrt{N}}\right),
  	\end{align*}
where the inequality $(\Delta_9)$ holds due to Theorem 3.3 of \cite{zhong2024neural}. Furthermore, the last equality holds by Assumptions (A\ref{assump: covariates}), (A\ref{assump: AsympNormal}) and (A\ref{assump:rate_lambda}).  
Moreover, a calculation yields $\Psi_0(\xi_0,h_0)=0$, so the left hand side of \eqref{proof: raw score} satisfies
  \begin{equation*}
  	\sqrt{N}\{\Psi_0(\hat{\xi}_N,\hat{h}_N)+\Psi_N(\xi_0,h_0)\}=o_p(1),
  \end{equation*}
  or equivalently,
  \begin{equation*}
  	\sqrt{N}\Psi_0(\hat{\xi}_N,\hat{h}_N)=-\sqrt{N}\Psi_N(\xi_0,h_0)+o_p(1).
  \end{equation*}

  Applying the Taylor's expansion for $\Psi_0(\xi,h)$ at $(\xi_0,h_0)$, we obtain
  \begin{align*}
  	&\Psi_0(\hat{\xi}_N,\hat{h}_N)\\
  	=&\mathbb{E}[-\mathbf{sign}(\varepsilon-\xi^\top_{(t)} \tilde{X}-h_{(t)}(Z))\tilde{X}]\Big|_{t=1}\\
  	=&\mathbb{E}_V[(2F_{\varepsilon}(\xi_{(t)}^\top\tilde{X}+h_{(t)}(Z)|V)-1)\tilde{X}]\Big|_{t=1}\\
  	=&\mathbb{E}[-\mathbf{sign}(\varepsilon)\tilde{X}]+\frac{\partial\mathbb{E}_V[(2F_{\varepsilon}(\xi_{(t)}^\top\tilde{X}+h_{(t)}(Z)|V)-1)\tilde{X}]}{\partial t}\Big|_{t=0}+O(d^2(\hat{\theta}_N,\theta_0)) \\
  	=&2\mathbb{E}_V\{f_{\varepsilon}(0|V)\tilde{X}\tilde{X}^\top\} (\hat{\xi}_N-\xi_0)+2\mathbb{E}_V\{f_{\varepsilon}(0|V)(\hat{h}_N-h_0)\tilde{X}^\top\} +O(d^2(\hat{\theta}_N,\theta_0))\\
  	=&2\mathbb{E}_V\{f_{\varepsilon}(0|V)\tilde{X}\tilde{X}^\top\} (\hat{\xi}_N-\xi_0)+2\mathbb{E}_V\{f_{\varepsilon}(0|V)(\hat{h}_N-h_0)(X-\boldsymbol{{\varphi}}^*(Z))^\top\}+O(d^2(\hat{\theta}_N,\theta_0))\\
  	=& 2\mathbb{E}_V\{f_{\varepsilon}(0|V)\tilde{X}\tilde{X}^\top\} (\hat{\xi}_N-\xi_0)+O(d^2(\hat{\theta}_N,\theta_0)).
  \end{align*}
Here the derivative w.r.t. $h$ and $\xi$ are based on the derivatives of the line $h_{(t)}=(1-t)h_0+t\hat{h}_N, t\in[0,1]$ and $\xi_{(t)}=(1-t)\xi_0+t\hat{\xi}_N, t\in[0,1]$ w.r.t. $t.$ Furthermore, the last equality holds by the orthogonality of $X-\boldsymbol{{\varphi}}^*(Z)$ w.r.t. $f(Z)\in L^2(P_Z)$ from the definition of $\boldsymbol{\varphi}^*(Z)$. Since $\hat{\xi}_N-\xi_0=\hat{\beta}_N-\beta_0$ and Assumption (A\ref{assump: AsympNormal}), it follows that
  \begin{equation*}
  	\sqrt{N}(\hat{\beta}_N-\beta_0)= \frac{1}{2}[\mathbb{E}_V\{f_{\varepsilon}(0|V)\tilde{X}\tilde{X}^\top\}]^{-1}\sqrt{N}\Psi_N(\xi_0,h_0)+o_{p}(1)\leadsto \mathcal{N}(0,\Sigma_2^{-1}\Sigma_1\Sigma_2^{-1}).
  \end{equation*}
  Therefore, the result follows.
\end{proof}

\section{Optimization Perspective}

\subsection{Continuous Approximation Approach}\label{approximate}

In this subsection, we design an efficient algorithm for solving optimization problem \eqref{opt2}, which can be rewritten into the following form:
\begin{equation}\label{optsec-1}
\begin{aligned}
	\min_{\beta, \mathbf{W}}~~& \frac{1}{N} \sum_{i=1}^{N} \left|Y_i - \beta^\top X_i - g(\mathbf{W};Z_i)\right|+\sum_{k=1}^{L}\delta_{\|W_k\|_{\infty}\le 1}(W_k)+\lambda_N \mathcal{J}_{N,M}(\beta,\mathbf{W}),\\
\mathrm{s.t.}~~&\sum_{k=1}^{L}\|W_k\|_{0}\le s,
\end{aligned}	
\end{equation}
where for any subset $C\subseteq\mathbb{R}^n$,
\begin{equation*}
  \delta_C(x) = \left\{
  \begin{array}{ll}
    0, & \text{if } x \in C, \\
    +\infty, & \text{if } x \notin C.
  \end{array}
  \right.
\end{equation*}
 Due to the introduction of coupled nonconvex constraint $\sum_{k=1}^{L}\|W_k\|_{0}\le s$, problem \eqref{optsec-1} is very challenging; hence we relax the abovementioned constraint into the following separable form:
\begin{equation}\label{optsec-3}
\begin{aligned}
	\min_{\beta, \mathbf{W}}~~\mathcal{L}_N(\beta,\mathbf{W};\mathbf{U})\doteq\underbrace {\frac{1}{N} \sum_{i=1}^{N} \left|Y_i - \beta^\top X_i - g(\mathbf{W};Z_i)\right|}_{\mathcal{R}_N(\beta, \mathbf{W}; \mathbf{U})}&+ \sum_{k=1}^{L}\left(\delta_{\|W_k\|_{\infty}\le 1}(W_k)+\gamma_k\Vert W_k\Vert_0  \right)\\
	&+\lambda_N {\mathcal{J}}_{N,M}(\beta, \mathbf{W})
\end{aligned}	
\end{equation}
 weighting parameter $\{\gamma_k\}_{k=1}^L$.

\begin{definition}
Let $f\colon\mathbb{R}\to \mathbb{R}$ and define $f_{\sigma}(y)\doteq f(y/\sigma)$ for any $\sigma>0$. The function $f$ is said to possess Property $\mathcal{D}$, if
\begin{enumerate}
  \item $f$ is real analytic on $(y_0,\infty)$ for some $y_0<0$,
  \item $\forall y\geq 0$, $f''(y)\geq -\mu_0$, where $\mu_0>0$ is some constant,
  \item $f$ is concave on $\mathbb{R}$,
  \item $f(y)=0\Leftrightarrow y=0$,
  \item $\lim_{y\to\infty}f(y)=1$.
\end{enumerate}
\end{definition}
It is obvious that if $f$ possesses Property $\mathcal{D}$, then
\begin{equation*}
  \lim\limits_{\sigma \downarrow 0^+} f_{\sigma}(|y|)=I(y)=\left\{
  \begin{aligned}
  &0,~~y=0,\\
  &1,~~\text{otherwise}.
  \end{aligned}
  \right.
\end{equation*}
In fact, there are a plenty of functions that satisfy Property $\mathcal{D}$, for instance, $f(y)=1-e^{-y}$. For $x=(x_1,\dots,x_n)^\top\in\mathbb{R}^n$, denote $f_{\sigma}(x)=\sum_{i=1}^{n}f_{\sigma}(x_i)$.
Hence, problem \eqref{optsec-3} may be approximated by the following continuous optimization problem
\begin{equation}\label{optsec-4}
\begin{aligned}
	\min_{\beta, \mathbf{W}}~~&\mathcal{L}_{N,\sigma}(\beta,\mathbf{W};\mathbf{U})\doteq\mathcal{R}_N(\beta, \mathbf{W}; \mathbf{U})+ \sum_{k=1}^{L}\left(\delta_{\|W_k\|_{\infty}\le 1}(W_k)+\gamma_k f_{\sigma}(|W_k|) \right)+\lambda_N {\mathcal{J}}_{N,M}(\beta,\mathbf{W}).
\end{aligned}	
\end{equation}

\begin{theorem}\label{thm_app}
  Let $\sigma_k\downarrow 0$, then the following statements hold.
  \begin{enumerate}
    \item $\inf\limits_{\beta,\mathbf{W}}\mathcal{L}_{N,\sigma_k}(\beta,\mathbf{W};\mathbf{U})\to\inf\limits_{\beta,\mathbf{W}}\mathcal{L}_{N}(\beta,\mathbf{W};\mathbf{U})$.
    \item For $v$ in some index set $N\in\mathcal{N}_{\infty}$, the sets ${\operatorname*{argmin}}~\mathcal{L}_{N,\sigma_k}$ are nonempty and form a bounded sequence with
        $$\limsup_k\left({\operatorname*{argmin}}~\mathcal{L}_{N,\sigma_k}\right)\subseteq{\operatorname*{argmin}}~ \mathcal{L}_{N}.$$
    \item For any choice of $\epsilon_k\downarrow 0$ and $(\beta_k,\mathbf{W}_k)\in\epsilon_k\text{-}{\operatorname*{argmin}}~\mathcal{L}_{N,\sigma_k}$, the sequence $\{(\beta_k,\mathbf{W}_k)\}_{k\in\mathbb{N}}$ is bounded and such that all its cluster points belong to ${\operatorname*{argmin}}~\mathcal{L}_N$.
  \end{enumerate}
\end{theorem}

\begin{proof}
Since for $x=(x_1,\dots,x_n)^\top \in\mathbb{R}^n$
\begin{equation*}
f_{\sigma}(x)=\sum_{i=1}^{n}f_{\sigma}(x_i),
\end{equation*}
we have $f_{\sigma_{k+1}}(x)\geq f_{\sigma_k}(x)$ for every $x\geq 0$. Hence, for every $(\beta,\mathbf{W})$, it follows that $\mathcal{L}_{N,\sigma_{k+1}}(\beta,\mathbf{W};\mathbf{U})\geq \mathcal{L}_{N,\sigma_k}(\beta,\mathbf{W};\mathbf{U})$, and $\{\mathcal{L}_{N,\sigma_k}(\beta,\mathbf{W};\mathbf{U})\}_{k\in\mathbb{N}}$ is nondecreasing. By Proposition 7.4 of \cite{Rockafellar1998}, ${\operatorname*{e-lim}}_k \mathcal{L}_{N,\sigma_k}$ exists and equals $\sup_k[{\operatorname*{cl}}\mathcal{L}_{N,\sigma_k}]$. Based on the fact  $\lim\limits_{\sigma\downarrow 0^+}f_{\sigma}(|x|)=\Vert x\Vert_0$, it follows that $\sup_k[{\operatorname*{cl}}\mathcal{L}_{N,\sigma_k}](\beta,\mathbf{W};\mathbf{U})=\mathcal{L}_{N}(\beta,\mathbf{W};\mathbf{U})$. Obviously, for every $\sigma_k$, $\mathcal{L}_{N,\sigma_k}(\beta,\mathbf{W};\mathbf{U})$ is a coercive function.
 According to Exercise 7.32 of \cite{Rockafellar1998}, the sequence $\{\mathcal{L}_{N,\sigma_k}\}_{k\in\mathbb{N}}$ is eventually level-bounded. By noticing that $\mathcal{L}_{N,\sigma_k}$ and $\mathcal{L}_{N}$ are l.s.c. and proper, we finish the proof by Theorem 7.33 of \cite{Rockafellar1998}.
\end{proof}

Problem \eqref{optsec-4} can be rewritten into the following form with bounded feasible set:
\begin{equation}\label{optsec-5}
\begin{aligned}
	\min_{\beta, \mathbf{W}}~~&{G}(\beta, \mathbf{W}; \mathbf{U})\doteq\mathcal{R}_N(\beta, \mathbf{W}; \mathbf{U})+ \sum_{k=1}^{L}\gamma_k f_{\sigma}(|W_k|) +\lambda_N {\mathcal{J}}_{N,M}(\beta, \mathbf{W})\\
	\mathrm{s.t.}~~& W_k\in [-1,1]^{q_k\times (q_{k-1}+1)},~k=1,\dots,L,\\
	&\beta \in [-{C},{C}]^d.
\end{aligned}	
\end{equation}
We fix a probability space $(\Omega^\prime,\mathcal{F}^\prime, \mathbb{P}^\prime)$ and equip $\mathcal{X}$ with the Borel $\sigma$-algebra with
\begin{equation*}
	\mathcal{X}\doteq [-{C},{C}]^d \times \prod_{k=1}^L[-1,1]^{q_k\times (q_{k-1}+1)}.
\end{equation*}
We suppose that there exists a measurable mapping $\zeta\colon \mathcal{X}\times \Omega^\prime\to \mathbb{R}\times \prod_{k=1}^L \mathbb{R}^{q_k\times (q_{k-1}+1)}$ satisfying:
\begin{equation*}
	\mathbb{E}_{\omega^\prime}[\zeta(\beta,\mathbf{W},\omega^\prime)]\in \partial_{(\beta, \mathbf{W})}^{C}{G}(\beta, \mathbf{W}; \mathbf{U})~~\text{for all}~~(\beta,\mathbf{W})\in\mathcal{X}.
\end{equation*}
In this section, we aim to analysis the proximal stochastic subgradient method that performs the following update rule
\begin{equation}\label{proximal-subgradient}
\left\{
  \begin{aligned}
  &\text{Sample}~\omega_k^\prime \sim \mathbb{P}^\prime,\\
  &(\beta_{k+1}, \mathbf{W}_{k+1}^\top)^\top\in \mathbf{Proj}_\mathcal{X}((\beta_{k}, \mathbf{W}_{k}^\top)^\top-\alpha_k \zeta(\beta_{k}, \mathbf{W}_{k},\omega_k^\prime))
  \end{aligned}
\right.
\end{equation}
with given an iterate $(\beta_{k}, \mathbf{W}_{k})\in\mathcal{X}$.

\begin{assumption}\label{ass_2}
We assume the following assumptions hold.
  \begin{itemize}
    \item The sequence $\{\alpha_k\}_{k\geq 1}$ is nonnegative, square summable, but not summable:
        \begin{equation*}
          \alpha_k\geq0,~~\sum_{k=1}^{\infty} \alpha_k=\infty,~~\text{and}~~\sum_{k=1}^{\infty} \alpha_k^2=\infty.
        \end{equation*}
    \item There exists a function $p\colon \mathcal{X}\to \mathbb{R}_+$, that is bounded on bounded sets, such that
        \begin{equation*}
          \mathbb{E}_{\omega^\prime}[\Vert\zeta(\beta,\mathbf{W},\omega^\prime)\Vert^2]\leq p(\beta, \mathbf{W})~~\text{for all}~~(\beta,\mathbf{W})\in\mathcal{X}.
        \end{equation*}
    \item For every convergent sequence $\{z_k\}_{k\geq 1}$, we have
    \begin{equation*}
      \mathbb{E}_{\omega^\prime}\left[\sup\limits_{k\geq 1} \Vert\zeta(\beta_k,\mathbf{W}_k,\omega^\prime)\Vert \right]<\infty.
    \end{equation*}
  \end{itemize}
\end{assumption}

\begin{theorem}\label{thm4}
  Let $\{(\beta_k,\mathbf{W}_k)\}_{k\geq 1}$ be the iterates produced by the proximal stochastic subgradient method \eqref{proximal-subgradient}. The almost surely, for all $(\beta^*,\mathbf{W}^*)\in\mathbf{Cluster}(\{(\beta_k,\mathbf{W}_k)\}_{k\geq 1})$, it holds that
  \begin{equation*}
    0 \in \partial_{(\beta, \mathbf{W})}^C{G}(\beta^*, \mathbf{W}^*; \mathbf{U})+ N_{\mathcal{X}}(\beta^*, \mathbf{W}^*),
  \end{equation*}
and the function values $\{{G}(\beta_k, \mathbf{W}_k; \mathbf{U})\}_{k\geq 1}$ converge.
\end{theorem}

\begin{proof}
To prove this theorem, by Theorem 6.2 of \cite{davis2020stochastic} and Assumption \ref{ass_2}, it suffices to show the descent property and weak Sard property. By Example 2.4 of \cite{kranz2025sad}, ${G}$ is definable in an o-minimal structure. Then, by Theorem 5.8 of \cite{davis2020stochastic}, ${G}$ and $1_\cX$ admit the chain rule. Therefore, the descent property holds by Lemma 6.3 of \cite{davis2020stochastic}. Thus we only argue the weak Sard property. Since ${G}$, and $1_\cX$ are definable in an o-minimial structure, there exist Whitney $C^{d+H}$-stratifications $\mathcal{A}_{G}$, and $\mathcal{A}_{\cX}$  of $\mathbf{graph}(G)$,  and $\cX$, respectively with $H=\prod_{k=1}^{L}(q_k+1)$. Let $\mathbf{Proj}(\mathcal{A}_{G})$  be the Whitney stratifications of $\R^{d+H}$ obtained by applying the coordinate projection $(\beta, \mathbf{W},r)\mapsto (\beta, \mathbf{W})$ to each stratum in $\mathcal{A}_{G}$. Appealing to Theorem 4.8 of \cite{van1996geometric}, we obtain a Whitney $C^{d+H}$-stratification $\mathcal{A}$ of $\R^{d+H}$ such that for every strata $M\in \mathcal{A}$ and $L\in \mathbf{Proj}(\mathcal{A}_{G})\cup \mathcal{A}_\cX$, either $M\cap L=\emptyset$ or $M\subseteq L$.

Consider an arbitrary stratum $M\in\mathcal{A}$ with  $M\cap \cX\neq \emptyset$ and a point $x\in M$. Obviously, we have $M\subseteq \cX$. Select the unique strata $M_{G}\in \mathbf{Proj}(\mathcal{A}_{{G}})$, and $M_{\cX}\in \mathcal{A}_\cX$ containing $x$. Let $\widehat{{G}}$  be $C^{d+H}$-smooth functions agreeing with ${G}$  on a neighborhood of $x$ in $ M_{G}$. By Proposition 4 of \cite{bolte2007clarke}, we conclude
$$\partial^C {G}(\beta,\mathbf{W})\subseteq \nabla \widehat{{G}}(\beta,\mathbf{W})+N_{ M_{G}}(\beta,\mathbf{W}),~ N_{\cX}(\beta,\mathbf{W})\subseteq N_{M_{\cX}}(\beta,\mathbf{W}).$$
Hence summing yields
\begin{align*}
\partial^C{G}(\beta,\mathbf{W})+N_{\cX}(\beta,\mathbf{W})&\subseteq \nabla (\widehat{{G}})(\beta,\mathbf{W})+N_{ M_{G}}(\beta,\mathbf{W})+N_{M_{\cX}}(\beta,\mathbf{W})\\
&\subseteq \nabla (\widehat{ {G}})(\beta,\mathbf{W})+N_{M}(\beta,\mathbf{W}),
\end{align*}
where the last inclusion follows from $M\subseteq M_{G}$ and $M\subseteq M_{\mathcal{X}}$ .
Notice that $\widehat{{G}}$ agrees with ${G}$ on a neighborhood of $(\beta,\mathbf{W})$ in $M$. Hence if the inclusion, $0\in \partial^C {G}(\beta,\mathbf{W})+N_{\cX}(\beta,\mathbf{W})$, holds it must be that $(\beta,\mathbf{W})$ is a critical point of the $C^{d+H}$-smooth function ${G}$ restricted to $M$, in the classical sense. Applying the Theorem 6.10 (classical Sard's theorem) of \cite{lee2003smooth}  to each manifold $M$, weak Sard's property holds. Hence, we finish the proof by Theorem 6.2 of \cite{davis2020stochastic}.
\end{proof}

\begin{remark}
The core idea of the proof above is to establish the Weak Sard property, which allows us to follow the line of \cite{davis2020stochastic} to complete the proof.
First, we partition the feasible set $\mathcal{X}$ into a collection of disjoint smooth manifolds $\mathcal{A}$, such that objective function \eqref{optsec-5} is smooth on each manifold $M \in \mathcal{A}$. We then apply the projection theorem (Proposition 4 of \cite{bolte2007clarke}) to show $\partial^C G(\beta, \mathbf{W})\subseteq\nabla \widehat{G}(\beta,\mathbf{W}) + N_{M_G}(\beta,\mathbf{W})$ and $N_{\mathcal{X}}(\beta,\mathbf{W}) \subseteq N_{M_\mathcal{X}}(\beta,\mathbf{W})$. Subsequently, we demonstrate that the summation of the classical gradient of the local mollifier $\nabla \widehat{G}(\beta,\mathbf{W})$ and the normal cone $N_M(\beta,\mathbf{W})$ covers  $\partial^C{G}(\beta,\mathbf{W}) + N_{\cX}(\beta,\mathbf{W})$. Finally, the Weak Sard Property is deduced by applying the standard Sard's theorem to the local mollifier $\widehat{G}(\beta,\mathbf{W})$ on each manifold in the partition.
\end{remark}
\subsection{Non-Approximation Approach}\label{non-approximat}

In this subsection, we directly solve the following optimization problem induced by $\ell_0$-norm:
\begin{equation}\label{direct1}
\begin{aligned}
	\min_{\beta, \mathbf{W}}~~& \mathcal{H}(\beta, \mathbf{W};\mathbf{U})\doteq\frac{1}{N} \sum_{i=1}^{N} \left|Y_i - \beta^\top X_i - g(\mathbf{W};Z_i)\right|+\lambda_N \mathcal{J}_{N,M}(\beta,\mathbf{W}),\\
\mathrm{s.t.}~~&\beta \in [-{C},{C}]^d,\\
&\mathbf{W}\in \mathcal{W}\doteq \left\{\mathbf{W}\colon \sum_{k=1}^{L}\|W_k\|_{0}\le s\right\} \bigcap \prod_{k=1}^L[-1,1]^{q_k\times (q_{k-1}+1)}.
\end{aligned}	
\end{equation}
Again, we fix a probability space $(\Omega^{\prime\prime},\mathcal{F}^{\prime\prime}, \mathbb{P}^{\prime\prime})$ and equip $[-{C},{C}]^d\times \mathcal{W}$ with the Borel $\sigma$-algebra.
We suppose that there exists a measurable mapping $\widetilde{\zeta}\colon [-{C},{C}]^d\times \mathcal{W}\times \Omega^{\prime\prime}\to \mathbb{R}\times \prod_{k=1}^L \mathbb{R}^{q_k\times (q_{k-1}+1)}$ satisfying:
\begin{equation*}
	\mathbb{E}_{\omega^{\prime\prime}}[\widetilde{\zeta}(\beta,\mathbf{W},\omega^\prime)]\in \partial^C_{(\beta, \mathbf{W})}{G}(\beta, \mathbf{W}; \mathbf{U})~~\text{for all}~~(\beta,\mathbf{W})\in[-{C},{C}]^d\times \mathcal{W}.
\end{equation*}
We still consider the proximal stochastic subgradient method that performs the following update rule
\begin{equation}\label{proximal-subgradient2}
\left\{
  \begin{aligned}
  &\text{Sample}~\omega_k^{\prime\prime} \sim \mathbb{P}^{\prime\prime},\\
  &\beta_{k+1} \in \mathbf{Proj}_{[-{C},{C}]^d}(\beta_k - \alpha_k \mathbf{Proj}_1\left(\widetilde{\zeta}(\beta_k,\mathbf{W}_k,\omega^{\prime\prime}_k)\right),\\
  &\mathbf{W}_{k+1} \in \mathbf{Proj}_{\mathcal{W}}(\mathbf{W}_k - \alpha_k \mathbf{Proj}_2\left(\widetilde{\zeta}(\beta_k,\mathbf{W}_k,\omega^{\prime\prime}_k)\right),
  \end{aligned}
\right.
\end{equation}
with given an iterate $(\beta_{k}, \mathbf{W}_{k})\in\mathcal{X}$.

\begin{lemma}
  The sub-routine
  \begin{equation}\label{sub1}
    \mathbf{W}^*\in\mathbf{Proj}_{\mathcal{W}} \left(\mathbf{W} - \alpha (\widetilde{\zeta}(\beta,\mathbf{W},\omega^{\prime\prime}_k))\right)
  \end{equation}
  admits the following closed-form solution. Denoting $G_{i}\doteq {W}_{i}-\alpha_k \mathbf{Proj}_2\left[ (\widetilde{\zeta}(\beta,\mathbf{W},\omega^{\prime\prime}))\right]_i$, we compute local benefit values
  \begin{equation*}
    \Delta_{i;j,k}=\left\{
    \begin{aligned}
    &[G_i]_{jk}^2,~~~~~~~~~~|[G_i]_{jk}|\leq 1,\\
    &2|[G_i]_{jk}|-1,~~|[G_i]_{jk}|>1,
    \end{aligned}
    \right.
  \end{equation*}
  for each entry $(i,j,k)\in [L]\times [q_i]\times [q_{i-1}+1]$. Let $\mathbb{T}\subseteq [L]\times [q_i]\times [q_{i-1}+1]$ contains the indices of the $s$ largest values of $\Delta_{i;j,k}$ (ties may be broken arbitrarily). Then the projection $\mathbf{W}^*$ can be selected as
  \begin{equation*}
    [\mathbf{W}^*_{i}]_{jk} = \left\{
    \begin{aligned}
    &\mathbf{clip}([G_i]_{jk},-1,1),~~(i,j,k)\in\mathbb{T},\\
    &0,~~(i,j,k)\in\mathbb{T},
    \end{aligned}
    \right.
  \end{equation*}
  where $\mathbf{clip}(y,-1,1) = \min \{1, \max\{-1,y\}\}$.
\end{lemma}
\begin{proof}
The sub-routine \eqref{sub1} is equivalent to the following optimization problem
\begin{equation*}
  \begin{aligned}
  \min_{\mathbf{W}}~~& \sum_{i=1}^{L} \Vert W_i - G_i\Vert^2=\sum_{i=1}^{L}\sum_{j=1}^{q_i}\sum_{k=1}^{q_{i-1}+1}\vert [W_i]_{jk}-[G_i]_{jk}\vert^2\\
  \mathrm{s.t.}~~& \mathbf{W}\in \mathcal{W}
  \end{aligned}
\end{equation*}
with given $\mathbf{G}=(G_1,\dots, G_L)$. For the sake of simplicity, we denote objective function as
\begin{equation*}
  \sum_{i=1}^{L}\sum_{j=1}^{q_i}\sum_{k=1}^{q_{i-1}+1}\vert [W_i]_{jk}-[G_i]_{jk}\vert^2\doteq \sum_{(i,j,k)} \vert [W_i]_{jk}-[G_i]_{jk}\vert^2.
\end{equation*}
For any fixed set $\mathbb{T}^\prime\subseteq [L]\times [q_i]\times [q_{i-1}+1]$ with $|\mathbb{T}^\prime|\leq s$, we consider the restricted feasible set
\begin{equation*}
  \mathcal{W}_{\mathbb{T}^\prime}=\{\mathbf{W}\colon [W_{i}]_{ij}=0~\text{for}~(i,j,k)\notin \mathbb{T}^\prime,~| [W_{i}]_{ij}|\leq 1~\forall (i,j,k)\in\mathbb{T}^\prime\}.
\end{equation*}
Then
\begin{equation*}
\begin{aligned}
  \min_{\mathbf{W}\in \mathcal{W}_{\mathbb{T}^\prime}}\sum_{(i,j,k)} \vert [W_i]_{jk}-[G_i]_{jk}\vert^2 &= \sum_{(i,j,k)\in \mathbb{T}^\prime} \min_{|[W_i]_{jk}|\leq 1}([W_i]_{jk}-[G_i]_{jk})^2 + \sum_{(i,j,k)\notin \mathbb{T}^\prime} [G_i]_{jk}^2\\
  &= \sum_{(i,j,k)\in \mathbb{T}^\prime} ([W_i]_{jk}^*-[G_i]_{jk})^2 + \sum_{(i,j,k)\notin \mathbb{T}^\prime} [G_i]_{jk}^2\\
  &= \sum_{(i,j,k)} [G_i]_{jk}^2 - \sum_{(i,j,k)\in \mathbb{T}^\prime} \Delta_{i;j,k}
\end{aligned}
\end{equation*}
with $[W_i]_{jk}^*\doteq \mathbf{clip}([G_i]_{jk},-1,1)$. Consequently, it holds that
\begin{equation*}
  \min_{\mathbf{W}\in \mathcal{W}}\sum_{(i,j,k)} \vert [W_i]_{jk}-[G_i]_{jk}\vert^2 =  \sum_{(i,j,k)} [G_i]_{jk}^2 - \max_{\mathbb{T}^\prime\colon |\mathbb{T}^\prime|\leq s} \sum_{(i,j,k)\in \mathbb{T}^\prime} \Delta_{i;j,k}
\end{equation*}
To maximize $\sum_{(i,j,k)\in \mathbb{T}^\prime}\Delta_{i;j,k}$ subject to $|\mathbb{T}^\prime|\leq s$,
the optimal $\mathbb{T}$ consists of the indices corresponding to the $s$ largest $\Delta_{i;j,k}$ (ties broken arbitrarily). Here, we complete the proof.
\end{proof}

\begin{theorem}
  Let $\{(\beta_k,\mathbf{W}_k)\}_{k\geq 1}$ be the iterates produced by the proximal stochastic subgradient method \eqref{proximal-subgradient2}. The almost surely, for all $(\beta^*,\mathbf{W}^*)\in\mathbf{Cluster}(\{(\beta_k,\mathbf{W}_k)\}_{k\geq 1})$, it holds that
  \begin{equation*}
    0 \in \partial^C_{(\beta, \mathbf{W})}\mathcal{H}(\beta^*, \mathbf{W}^*; \mathbf{U})+ N_{[-{C},{C}]^d\times \mathcal{W}}(\beta^*, \mathbf{W}^*),
  \end{equation*}
and the function values $\{\mathcal{H}(\beta_k, \mathbf{W}_k; \mathbf{U})\}_{k\geq 1}$ converge.
\end{theorem}

\begin{proof}
We only need to prove $\mathcal{W}$ is semi-algebraic; the remaining argument is identical to the proof of Theorem \ref{thm4}.
Write all entries of $\mathbf{W}$ as a single vector
$
x = (x_1, \dots, x_N) \in \mathbb{R}^H,
$
where $N$ is the total number of scalar elements in $(W_1, \dots, W_L)$.
For any index set $T \subset \{1, \dots, N\}$, define
\[
A_T = \{ x \in \mathbb{R}^{H} : x_i = 0 \ \text{for all}\ i \notin T \}.
\]
Each $A_T$ is the zero set of finitely many polynomial equations $\{x_i = 0 : i \notin T\}$,
so $A_T$ is an {algebraic} (hence semi-algebraic) subset of $\mathbb{R}^N$.
The condition $\sum_{k=1}^L \|W_k\|_0 \le s$ is equivalent to saying that
the total number of nonzero coordinates of $x$ is at most $s$.
Hence,
\[
\mathcal{W}
= \bigcup_{\substack{T \subset \{1, \dots, N\} \\ |T| \le s}} A_T.
\]
This is a finite union since there are only finitely many subsets $T$ with $|T| \le s$.
A finite union of semi-algebraic sets is again semi-algebraic,
and each $A_T$ is semi-algebraic.
Therefore, $\mathcal{W}$ is a semi-algebraic subset of $\mathbb{R}^{H}$.
\end{proof}


\begin{remark}
Under some additional mild assumptions, almost surely, the sequence $\{\beta_k,\mathbf{W}_k\}_{k\geq 1}$ converges to a local minimum of $\mathcal{H}$, i.e., the proximal stochastic subgradient method \eqref{proximal-subgradient2} can escape active strict saddles and sharply repulsive critical points of $\mathcal{H}$; the readers may refer to \cite{schechtman2021stochastic} for the details.
\end{remark}

\begin{remark}
	In fact, both the continuous approximation approach (proposed in Section \ref{approximate}) and the non-approximation approach (proposed in Section \ref{non-approximat}) admit independent research interests. On the one hand, although there exists a gap between continuous relaxation problem \eqref{optsec-3} and primal penalized LAD problem \eqref{opt2}, proximal stochastic subgradient update \eqref{proximal-subgradient} for problem \eqref{optsec-3} is very cheap, as it only requires projections onto a boxed set.
	Additionally, by Theorem \ref{thm_app},  relaxation problem \eqref{optsec-3} can approximate penalized LAD problem \eqref{opt2} to arbitrary accuracy, thereby exhibiting independent interest beyond serving as a computational surrogate.
	On the other hand, the non-approximation approach aims to solve problem \eqref{direct1} which is completely equivalent to penalized LAD problem \eqref{opt2}, henceforth enjoying high statistical accuracy. Nevertheless, the proximal stochastic subgradient update \eqref{proximal-subgradient2} involves a sorting operation with complexity $O(H\log H)$, and is therefore relatively more computationally demanding and unsuitable for ultra large-scale networks. Overall, these two approaches illustrate a fundamental trade-off between computational efficiency and statistical fidelity.
\end{remark}

\bibliographystyle{plain}
\bibliography{cite.bib}
\end{document}